\documentclass[12pt]{article}
\usepackage{dirtree}
\usepackage[edges]{forest}
\usepackage{graphicx}
\usepackage{color}
\usepackage{epsfig}
\usepackage{graphicx}
\usepackage{subfigure}
\usepackage{amsmath}
\usepackage{amssymb}
\usepackage{longtable}
\usepackage[ruled]{algorithm}
\usepackage{algorithmic}
\usepackage{graphicx}
\usepackage{hyperref}
\usepackage{setspace}
\usepackage{fullpage}
\usepackage{mathtools}
\usepackage{float}
\usepackage{array}
\newcolumntype{L}[1]{>{\raggedright\let\newline\\\arraybackslash\hspace{0pt}}m{#1}}
\newcolumntype{C}[1]{>{\centering\let\newline\\\arraybackslash\hspace{0pt}}m{#1}}
\newcolumntype{R}[1]{>{\raggedleft\let\newline\\\arraybackslash\hspace{0pt}}m{#1}}

\begin{document}

\title{Deep Reinforcement Learning in Computer Vision: \\ A Comprehensive Survey
}

\author{Ngan Le$^{**}$ \and Vidhiwar Singh Rathour$^*$ \and Kashu Yamazaki$^*$ \and Khoa Luu \and Marios Savvides}


\maketitle

\begin{abstract}
Deep reinforcement learning augments the reinforcement learning framework and utilizes the powerful representation of deep neural networks. Recent works have demonstrated the remarkable successes of deep reinforcement learning in various domains including finance, medicine, healthcare, video games, robotics, and computer vision. In this work, we provide a detailed review of recent and state-of-the-art research advances of deep reinforcement learning in computer vision. We start with \emph{comprehending the theories} of deep learning, reinforcement learning, and deep reinforcement learning. We then \emph{propose a categorization} of deep reinforcement learning methodologies and \emph{discuss their advantages and limitations}. In particular, we divide deep reinforcement learning into \emph{seven main categories} according to their applications in computer vision, i.e. (i) landmark localization (ii) object detection; (iii) object tracking; (iv) registration on both 2D image and 3D image volumetric data (v) image segmentation; (vi) videos analysis; and (vii) other applications. Each of these categories is further analyzed with reinforcement learning techniques, network design, and performance. Moreover, we provide a comprehensive analysis of the existing publicly available datasets and examine source code availability. Finally, we present some open issues and discuss future research directions on deep reinforcement learning in computer vision.

\end{abstract}

\section{Introduction}


Reinforcement learning (RL) is a machine learning technique for learning a sequence of actions in an interactive environment by trial and error that maximizes the expected reward \cite{sutton2018reinforcement}. Deep Reinforcement Learning (DRL) is the combination of \textit{Reinforcement Learning} and \textit{Deep Learning} (DL) and it has become one of the most intriguing areas of artificial intelligence today. DRL can solve a wide range of complex real-world decision-making problems with human-like intelligence that were previously intractable. DRL was selected by \cite{rotman2013technology}, \cite{giles2017technology} as one of ten breakthrough techniques in 2013 and 2017, respectively. 

The past years have witnessed the rapid development of DRL thanks to its amazing achievement in solving challenging decision-making problems in the real world. DRL has been successfully applied into many domains including games, robotics, autonomous driving, healthcare, natural language processing, and computer vision. In contrast to supervised learning which requires large labeled training data, DRL samples training data from an environment. This opens up many machine learning applications where big labeled training data does not exist.

Far from supervised learning, DRL-based approaches focus on solving sequential decision-making problems. They aim at deciding, based on a set of experiences collected by interacting with the environment, the sequence of actions in an uncertain environment to achieve some targets. Different from supervised learning where the feedback is available after each system action, it is simply a scalar value that may be delayed in time in the DRL framework. For example, the success or failure of the entire system is reflected after a sequence of actions. Furthermore, the supervised learning model is updated based on the loss/error of the output and there is no mechanism to get the correct value when it is wrong. This is addressed by policy gradients in DRL by assigning gradients without a differentiable loss function. This aims at teaching a model to try things out randomly and learn to do correct things more.


Many survey papers in the field of DRL including \cite{arulkumaran2017brief} \cite{franccois2018introduction} \cite{yu2019reinforcement} have been introduced recently. While \cite{arulkumaran2017brief} covers central algorithms in DRL, \cite{franccois2018introduction} provides an introduction to DRL models, algorithms, and techniques, where particular focus is the aspects related to generalization and how DRL can be used for practical applications. Recently, \cite{yu2019reinforcement} introduces a survey, which discusses the broad applications of RL techniques in healthcare domains ranging from dynamic treatment regimes in chronic diseases and critical care, an automated medical diagnosis from both unstructured and structured clinical data, to many other control or scheduling domains that have infiltrated many aspects of a healthcare system. Different from the previous work, our survey focuses on how to implement DRL in various computer vision applications such as landmark detection, object detection, object tracking, image registration, image segmentation, and video analysis.

Our goal is to provide our readers good knowledge about the principle of RL/DRL and thorough coverage of the latest examples of how DRL is used for solving computer vision tasks. We structure the rest of the paper as follows: we first introduce fundamentals of Deep Learning (DL) in section \ref{sec:dl} including Multi-Layer Perceptron (MLP), Autoencoder, Deep Belief Network, Convolutional Neural Networks (CNNs), Recurrent Neural Networks (RNNs). Then, we present the theories of RL in section \ref{sec:RL}, which starts with the Markov Decision Process (MDP) and continues with value function and Q-function. In the end of section \ref{sec:RL}, we introduce various techniques in RL under two categories of model-based and model-free RL. Next, we introduce DRL in section \ref{sec:drl} with main techniques in both value-based methods, policy gradient methods, and actor-critic methods under model-based and model-free categories. The application of DRL in computer vision will then be introduced in sections \ref{sec:landmarkdetection}, \ref{sec:objectdetection}, \ref{sec:objecttracking}, \ref{sec:imageregistration}, \ref{sec:imagesegmentation}, \ref{sec:videoapplications}, \ref{sec:otherapplications} corresponding respectively to DRL in landmark detection, DRL in object detection, DRL in object tracking, DRL in image registration, DRL in image segmentation, DRL in video analysis and other applications of DRL. Each application category first starts with a problem introduction and then state-of-the-art approaches in the field are discussed and compared through a summary table. We are going to discuss some future perspectives in section \ref{sec:future} including challenges of DRL in computer vision and the recent advanced techniques.

\section{Introduction to Deep Learning}
\label{sec:dl}

\subsection{Multi-Layer Perceptron (MLP)}
\begin{figure}[!h]
    \centering
        \includegraphics[width=0.4\textwidth]{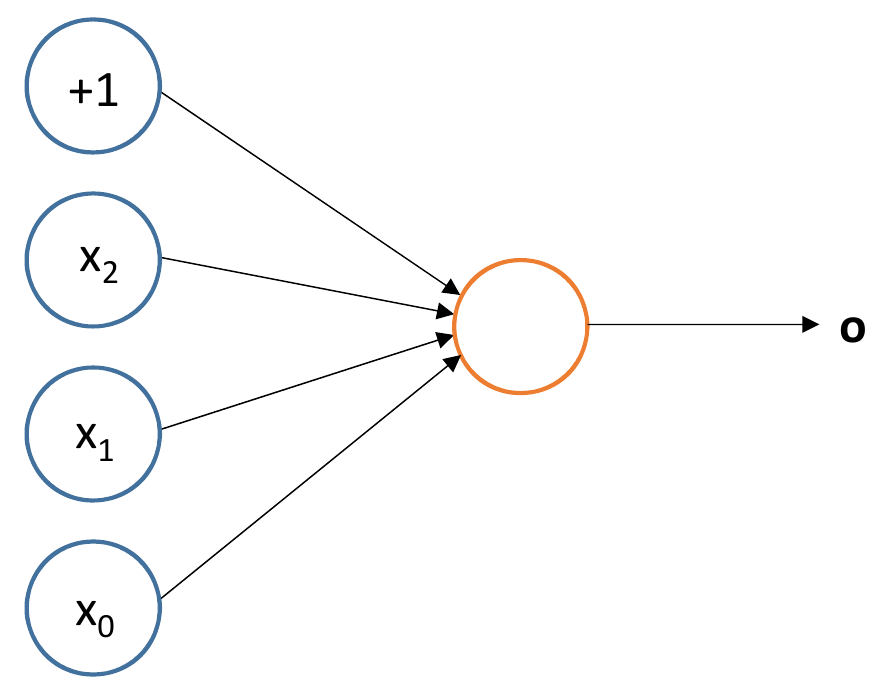}
        \caption{An example of one neuron which takes input $\textbf{x} = [x_1, x_2, x_3]$, the intercept term $+1$ as bias, and the output $\textbf{o}$.}
        \label{fig:neuron}
        
\end{figure}

Deep learning models, in simple words, are large and deep artificial neural networks. Let us consider the simplest possible neural network which is called "\textbf{neuron}" as illustrated in Fig. \ref{fig:neuron}. A computational model of a single neuron is called a perceptron which consists of one or more inputs, a processor, and a single output.

\begin{figure}[!h]
    \centering
        \includegraphics[width=0.4\textwidth]{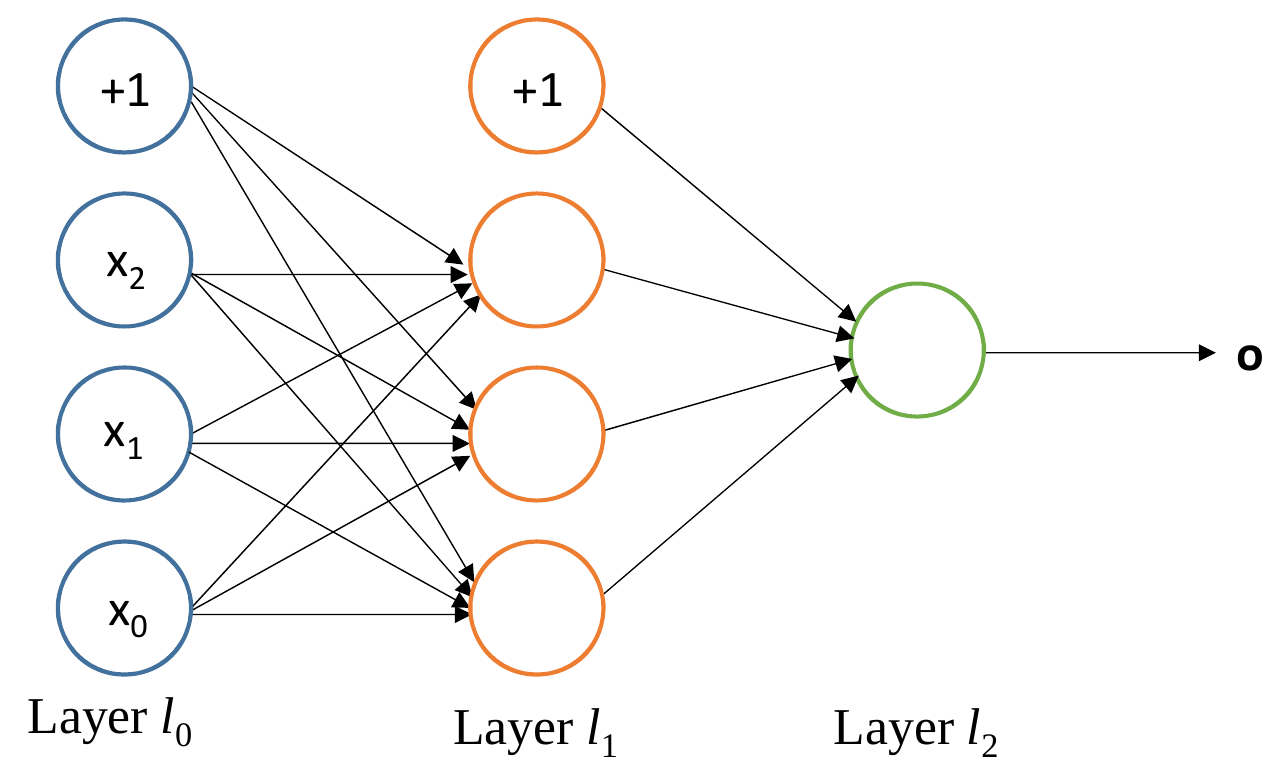}
        \caption{An example of multi-layer perceptron network (MLP)}
        \label{fig:active_func}
    
    \label{fig:CNN}
\end{figure}


\begin{figure*}[!]
    \centering
        \includegraphics[width=1.0\textwidth]{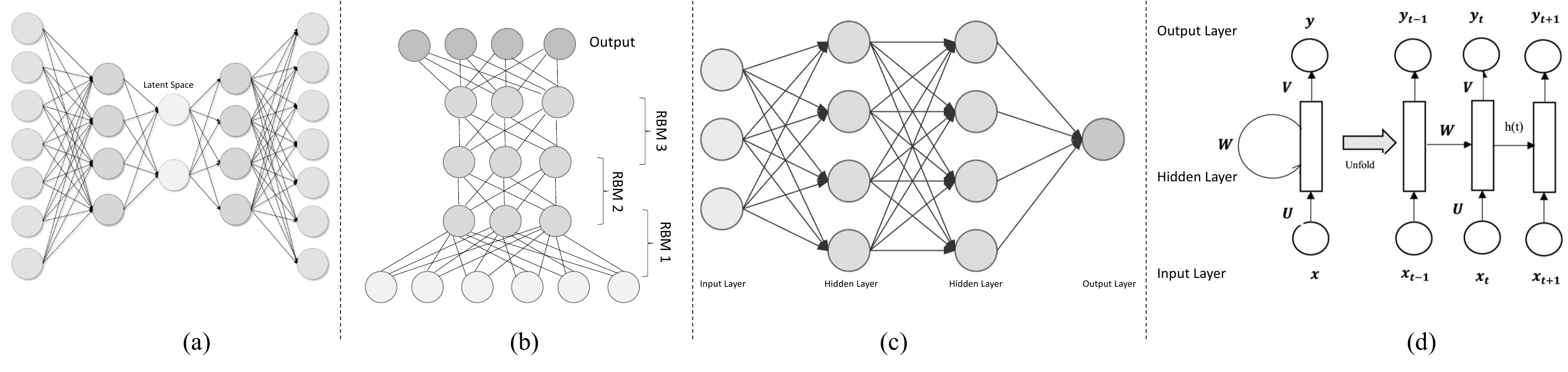}
        \caption{An illustration of various DL architectures. (a): Autoencoder (AE); (b): Deep Belief Network; (c): Convolutional Neural Network (CNN); (d): Recurrent Neural Network (RNN).}
        \label{fig:DL}
\end{figure*}



In this example, the neuron is a computational unit that takes $\textbf{x} = [x_0, x_1, x_2]$ as input, the intercept term $+1$ as bias $\textbf{b}$, and the output $\textbf{o}$. The goal of this simple network is to learn a function $f:\mathrm{R^N}\rightarrow \mathrm{R^M}$ where $N$ is the number of dimensions for input $\textbf{x}$ and $M$ is the number of dimensions for output which is computed as $\textbf{o} = f(\textbf{x}, \theta)$, where $\theta$ is a set of weights and are known as weights $\theta = \{w_i\}$. Mathematically, the output $\textbf{o}$ of a one neuron is defined as: 
\begin{equation}
\textbf{o} = f(\textbf{x}, \theta)
 = \sigma\left(\sum_{i=1}^{N}{w_ix_i + b}\right) 
 = \sigma(\textbf{W}^T\textbf{x} + b)
\end{equation}

In this equation, $\sigma$ is the point-wise non-linear activation function. The common non-linear activation functions for hidden units are hyperbolic tangent (\textit{Tanh}), sigmoid, softmax, ReLU, and LeakyReLU. A typical multi-layer perception (MLP) neural network is composed of one input layer, one output layer, and many hidden layers.  Each layer may contain many units. In this network, $\textbf{x}$ is the input layer, $\textbf{o}$ is the output layer. The middle layer is called the hidden layer. In Fig. \ref{fig:CNN}(b), MLP contains 3 units of the input layer, 3 units of the hidden layer, and 1 unit of the output layer. 

In general, we consider a MLP neural network with $L$ hidden layers of units, one layer of input units and one layer of output units. The number of input units is $N$, output units is $M$, and  units in hidden layer $l^{th}$ is $N^l$. The weight of the $j^{th}$ unit in layer $l^{th}$ and the $i^{th}$ unit in layer $(l+1)^{th}$ is denoted by $w_{ij}^{l}$. The activation of the $i^{th}$ unit in layer $l^{th}$ is $\textbf{h}_{i}^{l}$.

\begin{figure}[!h]
	\centering \includegraphics[width=0.8\columnwidth]{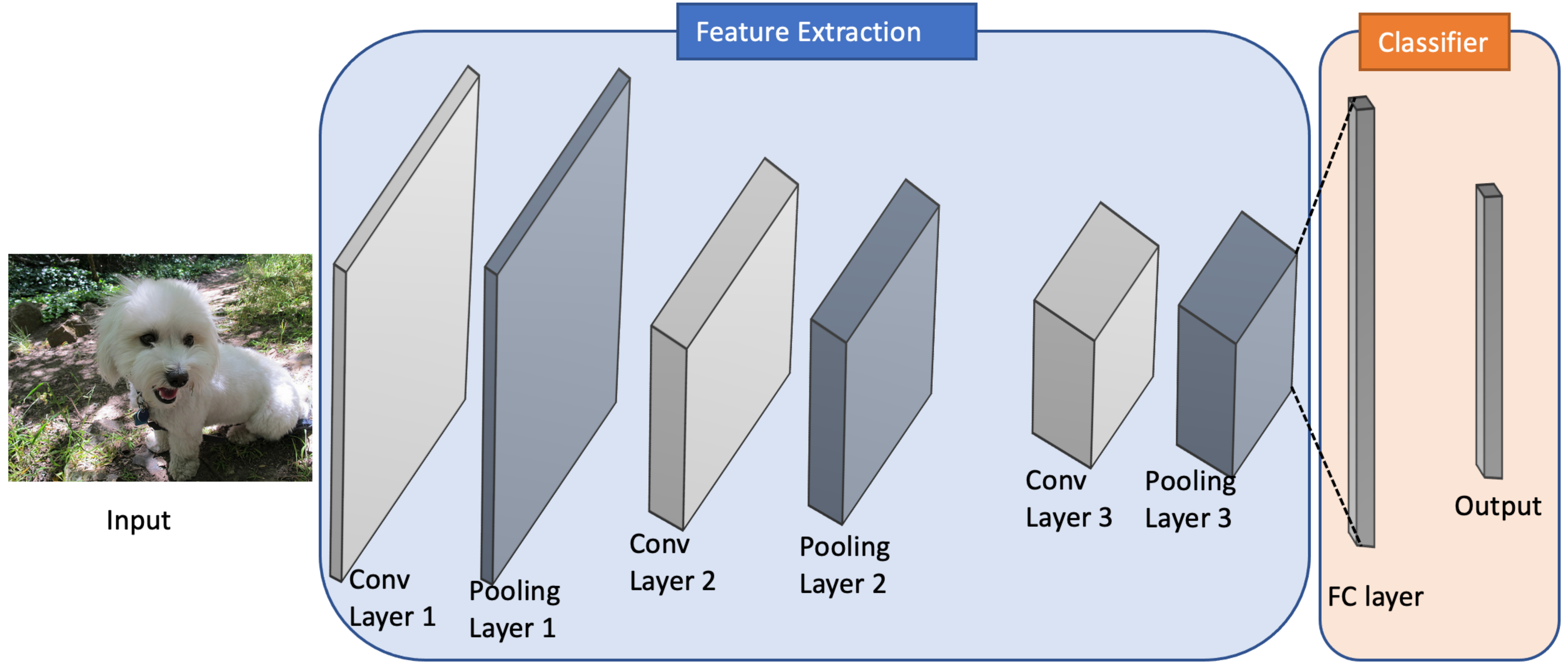}
		\caption[Architecture of a typical convolutional network for image classification]{Architecture of a typical convolutional network for image classification containing three basic layers: convolution layer, pooling layer and fully connected layer}
	\label{fig:basic_CNN}
\end{figure}

\subsection{Autoencoder}
Autoencoder is an unsupervised algorithm used for representation learning, such as feature selection or dimension reduction. A gentle introduction to Variational Autoencoder (VAE) is given in \cite{an2015variational} and VAE framework is illustrated in Fig.\ref{fig:DL}(a). In general, VAE aims to learn a parametric latent variable model by maximizing the marginal log-likelihood of the training data.

\subsection{Deep Belief Network}
Deep Belief Network (DBN) and Deep Autoencoder are two common unsupervised approaches that have been used to initialize the network instead of random initialization. While Deep Autoencoder is based on Autoencoder, Deep Belief Networks is based on Restricted Boltzmann Machine (RBM), which contains a layer of input data and a layer of hidden units that learn to represent features that capture high-order correlations in the data as illustrated in Fig.\ref{fig:DL}(b).

\subsection{Convolutional Neural Networks (CNN)}
Convolutional Neural Network (CNN) \cite{lecun1988theoretical} \cite{lecun1998efficient} is a special case of fully connected MLP that implements weight sharing for processing data. CNN uses the spatial correlation of the signal to utilize the architecture in a more sensible way. Their architecture, somewhat inspired by the biological visual system, possesses two key properties that make them extremely useful for image applications: spatially shared weights and spatial pooling. These kinds of networks learn features that are shift-invariant, i.e., filters that are useful across the entire image (due to the fact that image statistics are stationary). The pooling layers are responsible for reducing the sensitivity of the output to slight input shifts and distortions, and increasing the reception field for next layers. Since 2012, one of the most notable results in Deep Learning is the use of CNN to obtain a remarkable improvement in object recognition in ImageNet classification challenge \cite{deng2009imagenet} \cite{krizhevsky2012imagenet}.

A typical CNN is composed of multiple stages, as shown in Fig. \ref{fig:DL}(c). The output of each stage is made of a set of 2D arrays called feature maps. Each feature map is the outcome of one convolutional (and an optional pooling) filter applied over the full image. A point-wise non-linear activation function is applied after each convolution. In its more general form, a CNN can be written as
\begin{equation}
\begin{split}
\textbf{h}^0 = & \textbf{x}\\
\textbf{h}^l = & pool^l(\sigma_l(\textbf{w}^l\textbf{h}^{l-1} + \textbf{b}^l)), \forall l \in {1, 2, ... L}\\
\textbf{o} = & \textbf{h}^L\\
\end{split}
\end{equation}
where $\textbf{w}^l, \textbf{b}^l$ are trainable parameters as in MLPs at layer $l^{th}$. $\textbf{x}\in \mathrm{R}^{c\times h\times w}$ is vectorized from an input image with $c$ being the color channels, $h$ the image height and $w$ the image width. $\textbf{o}\in \mathrm{R}^{n\times h'\times w'}$ is vectorized from an array of dimension $h'\times w'$ of output vector (of dimension $n$). $pool^l$ is a (optional) pooling function at layer $l^{th}$.

Compared to traditional machine learning methods, CNN has achieved state-of-the-art performance in many domains including image understanding, video analysis and audio/speech recognition. In \textit{image understanding} \cite{xie2020self}, \cite{ZHAO2019251}, CNN outperforms human capacities \cite{BUETTIDINH2019e00321}. \textit{Video analysis} \cite{zhang2020multimodal}, \cite{LI2018323} is another application that turns the CNN model from a detector \cite{khoa_2021} into a tracker \cite{fan2010human}. As a special case of \textit{image segmentation} \cite{le2018reformulating}, \cite{le2018deep}, \textit{saliency detection} is another computer vision application that uses CNN \cite{wang2015deep}, \cite{li2015visual}. In addition to the previous applications, \textit{pose estimation} \cite{patacchiola2017head}, \cite{toshev2014deeppose} is another interesting research that uses CNN to estimate human-body pose. \textit{Action recognition} in both still images and videos is a special case of recognition and is a challenging problem. \cite{gkioxari2015contextual} utilizes CNN-based representation of contextual information in which the most representative secondary region within a large number of object proposal regions, together with the contextual features, is used to describe the primary region. CNN-based action recognition in video sequences is reviewed in \cite{zhang2016rgb}. \textit{Text detection and recognition} using CNN is the next step of optical character recognition (OCR) \cite{xu2015robust} and word spotting \cite{jaderberg2014deep}. Not only in computer vision, CNN has been successfully applied into other domains such as \textit{speech recognition and speech synthesis} \cite{8632885}, \cite{app9194050}, biometrics \cite{luu2016deep}, \cite{duong2019mobiface}, \cite{nhan2017temporal}, \cite{3190618},\cite{rathour2021roughness}, \cite{191200271}, biomedical \cite{le2020offset}, \cite{singh20203d}, \cite{le2020multi}, \cite{yamazaki2021invertible}.

\subsection{Recurrent Neural Networks (RNN)}
RNN is an extremely powerful sequence model and was introduced in the early 1990s \cite{jordan1990long}. A typical RNN contains three parts, namely, sequential input data ($\textbf{x}_t$), hidden  state ($\textbf{h}_t$) and sequential output data ($\textbf{y}_t$) as shown in Fig. \ref{fig:DL}(d). 

RNN makes use of sequential information and performs the same task for every element of a sequence where the output is dependent on the previous computations. The activation of the hidden states at time-step $t$ is computed as a function $f$ of the current input symbol $\bf{x}_t$ and the previous hidden states $\bf{h}_{t - 1}$. The output at time $t$ is calculated as a function $g$ of the current hidden state $\bf{h}_t$ as follows
\begin{equation}
\begin{split}
\textbf{\textbf{h}}_{t} & = f(\textbf{Ux}_t + \textbf{Wh}_{t-1}) \\
\textbf{y}_t & = g(\textbf{Vh}_t)
\end{split}
\label{eq:rnn}
\end{equation}

\noindent where $\bf{U}$ is the input-to-hidden weight matrix, $\bf{W}$ is the state-to-state recurrent weight matrix, $\bf{V}$ is the hidden-to-output weight matrix.  $f$ is usually a logistic sigmoid function or a hyperbolic tangent function and $g$ is defined as a softmax function.

Most works on RNN have made use of the method of backpropagation through time (BPTT) \cite{Rumelhart1998architecture} to train the parameter set $(\bf{U}, \bf{V}, \bf{W})$ and propagate error backward through time. In classic backpropagation, the error or loss function is defined as
\begin{equation}
E(\textbf{y'}, \textbf{y}) = \sum_t{||\textbf{y'}_t - \textbf{y}_t||^2}
\end{equation}
where $\textbf{y}_t$ is the prediction and $\textbf{y'}_t$ is the labeled groundtruth.

For a specific weight $\textbf{W}$, the update rule for gradient descent is defined as $\textbf{W}^{new} = \textbf{W} - \gamma\frac{\partial E}{\partial \textbf{W}}$, where $\gamma$ is the learning rate. In RNN model, the gradients of the error with respect to our parameters $\textbf{U}$, $\textbf{V}$ and $\textbf{W}$ are learned using Stochastic Gradient Descent (SGD) and chain rule of differentiation.

The difficulty of training RNN to capture long-term dependencies has been studied in \cite{Bengio1994}. To address the issue of learning long-term dependencies, Hochreiter and Schmidhuber \cite{hochreiter1997long} proposed Long Short-Term Memory (LSTM), which can maintain a separate memory cell inside it that updates and exposes its content only when deemed necessary. Recently, a Gated Recurrent Unit (GRU) was proposed by \cite{Cho2014} to make each recurrent unit adaptively capture dependencies of different time scales. Like the LSTM unit, the GRU has gating units that modulate the flow of information inside the unit but without having separate memory cells. 

Several variants of RNN have been later introduced and successfully applied to wide variety of tasks, such as natural language processing \cite{MikolovKBCK2011}, \cite{Li2015}, speech recognition \cite{Graves2013}, \cite{Chorowski2015}, machine translation \cite{kalchbrenner2013}, \cite{Luong2014},  question answering \cite{Hill2015}, image captioning \cite{Mao2014}, \cite{Donahue2014}, and many more.

\section{Basics of Reinforcement Learning}
\label{sec:RL}
This section serves as a brief introduction to the theoretical models and techniques in RL. In order to provide a quick overview of what constitutes the main components of RL methods, some fundamental concepts and major theoretical problems are also clarified. 
RL is a kind of machine learning method where agents learn the optimal policy by trial and error. Unlike supervised learning, the feedback is available after each system action, it is simply a scalar value that may be delayed in time in RL framework, for example, the success or failure of the entire system is reflected after a sequence of actions. Furthermore, the supervised learning model is updated based on the loss/error of the output and there is no mechanism to get the correct value when it is wrong. This is addressed by policy gradients in RL by assigning gradients without a differentiable loss function which aims at teaching a model to try things out randomly and learn to do correct things more. 

Inspired by behavioral psychology, RL was proposed to address the sequential decision-making problems which exist in many applications such as games, robotics, healthcare, smart grids, stock, autonomous driving, etc. Unlike supervised learning where the data is given, an artificial agent collects experiences (data) by interacting with its environment in RL framework. Such experience is then gathered to optimize the cumulative rewards/utilities.

In this section, we focus on how the RL problem can be formalized as an agent that can make decisions in an environment to optimize some objectives presented under reward functions. Some key aspects of RL are: (i) Address the sequential decision making; (ii) There is no supervisor, only a reward presented as scalar number; and (iii) The feedback is highly delayed. Markov Decision Process (MDP) is a framework that has commonly been used to solve most RL problems with discrete actions, thus we will first discuss MDP in this section. We then introduce value function and how to categorize RL into model-based or model-free methods. At the end of this section, we discuss some challenges in RL. 
\begin{figure}[!h]
	\centering \includegraphics[width=0.5\columnwidth]{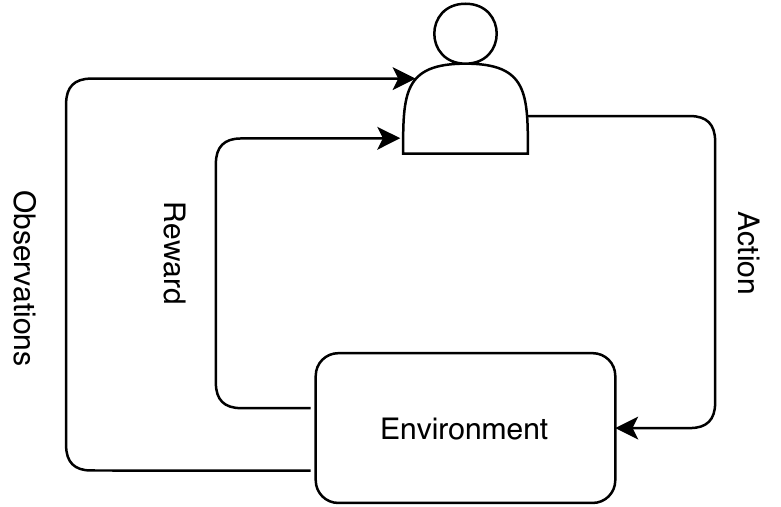}
	\caption{An illustration of agent-environment interaction in RL}
	\label{fig:RL}
\end{figure}



\subsection{Markov Decision Process}
The standard theory of RL is defined by a Markov Decision Process (MDP), which is an extension of the Markov process (also known as the Markov chain). Mathematically, the Markov process is a discrete-time stochastic process whose conditional probability distribution of the future states only depends upon the present state and it provides a framework to model decision-making situations. An MDP is typically defined by five elements as follows:
\begin{itemize}
    \item $S$: a set of \textit{state} or observation space of an environment. $s_0$ is starting state.
    \item $\mathcal{A}$: set of \textit{actions} the agent can choose.
    \item $T$: a \textit{transition probability} function $T(s_{t+1}| s_t, a_t)$, specifying the probability that the environment will transition to state $s_{t+1}\in S$ if the agent takes action $a_t \in \mathcal{A}$ in state $s_t \in S$.
    \item $R$:  a \textit{reward} function where $r_{t+1} = R(s_t, s_{t+1})$ is a reward received for taking action $a_t$ at state $s_t$ and transfer to the next state $s_{t+1}$.
    \item $\gamma$: a discount factor.
\end{itemize}

Considering MDP($S$, $\mathcal{A}$, $\gamma$, $T$, $R$), the agent chooses an action $a_t$ according to the policy $\pi(a_t|s_t)$ at state $s_t$. Notably, agent’s algorithm for choosing action $a$ given current state $s$, which in general can be viewed as distribution $\pi(a | s)$, is called a \textit{policy} (strategy). The environment receives the action, produces a reward $r_{t+1}$ and transfers to the next state $s_{t+1}$ according to the transition probability $T(s_{t+1}|s_t, a_t)$. The process continues until the agent reaches a terminal state or a maximum time step. In RL framework, the tuple $(s_t, a_t, r_{t+1}, s_{t+1})$ is called \textit{transition}. Several sequential transitions are usually referred to as roll-out. Full sequence $(s_0, a_0, r_1, s_1, a_1, r_2, ... )$ is called a \textit{trajectory}. Theoretically,  trajectory is infinitely long, but the episodic property holds in most practical cases. One trajectory of some finite length $\tau$ is called an \textit{episode}. For given MDP and policy $\pi$, the probability of observing $(s_0, a_0, r_1, s_1, a_1, r_2, ... )$  is called \textit{trajectory distribution} and is denoted as:
\begin{equation}
    \mathcal{T}_{\pi} = \prod_{t}{\pi(a_t|s_t)T(s_{t+1}|s_t, a_t)}
    \label{eq:T0}
\end{equation}


\noindent The objective of RL is to find the \textit{optimal policy} $\pi^*$ for the agent that maximizes the cumulative reward, which is called \textit{return}. For every episode, the return is defined as the weighted sum of immediate rewards:
\begin{equation}
 \mathcal{R} = \sum_{t=0}^{\tau-1}{\gamma^t r_{t+1}}
 \label{eq:R}
\end{equation}

\noindent Because the policy induces a trajectory distribution, the \textit{expected reward} maximization can be written as:
\begin{equation}
\mathbb{E}_{{\mathcal{T}_\pi}} \sum_{t=0}^{\tau-1}{r_{t+1}} \rightarrow \max_{\pi}
\end{equation}

\noindent Thus, given MDP and policy $\pi$, the \textit{discounted expected reward} is defined: 
\begin{equation}
    \mathcal{G}(\pi) = \mathbb{E}_{{\mathcal{T}_\pi}}\sum_{t=0}^{\tau-1}\gamma^{t}{r_{t+1}}
\end{equation}
The goal of RL is to find an \textit{optimal policy} $\pi^*$, which maximizes the discounted expected reward, i.e. $\mathcal{G}(\pi) \rightarrow \max_{\pi}$.

\subsection{Value and Q- functions}
The value function is applied to evaluate how good it is for an agent to utilize policy $\pi$ to visit state $s$. The concept of "good" is defined in terms of expected return, i.e. future rewards that can be expected to receive in the future and it depends on what actions it will take. Mathematically, the value is the expectation of return, and value approximation is obtained by Bellman expectation equation as follows:
\begin{equation}
\begin{split}
    V^\pi(s_t) = \mathbb{E}[r_{t+1} + \gamma V^\pi(s_{t+1})]
\end{split}    
\end{equation}
$V^\pi(s_t)$ is also known as state-value function, and the expectation term can be expanded as a product of policy, transition probability, and return as follows:
\begin{equation}
\begin{split}
    V^\pi(s_t) = \sum_{a_t\in\mathcal{A}}{\pi(a_t|s_t)}  \sum_{s_{t+1}\in 
    S}{T(s_{t+1} | s_t, a_t)[R(s_t, s_{t+1}) + \gamma V^\pi(s_{t+1})]}
\end{split} 
\label{eq:value}
\end{equation}
This equation is called the Bellman equation. When the agent always selects the action according to the optimal policy $\pi^*$ that maximizes the value, the Bellman equation can be expressed as follows:
\begin{equation}
\begin{split}
    V^*(s_t) = & \max_{a_t}\sum_{s_{t+1}\in S}{T(s_{t+1} | s_t, a_t)[R(s_t, s_{t+1}) + \gamma V^*(s_{t+1})]} \\ \overset{\Delta}{=} & \max_{a_t}Q^*(s_t, a_t)
\end{split}
\label{eq:bestvalue}
\end{equation}
However, obtaining optimal value function $ V^*$ does not provide enough information to reconstruct some optimal policy $\pi^*$ because the real-world environment is complicated. Thus, a quality function (Q-function) is also called the action-value function under policy $\pi$. The Q-function is used to estimate how good it is for an agent to perform a particular action ($a_t$) in a state ($s_t$) with a policy $\pi$ and it is introduced as:
\begin{equation}
    Q^\pi(s_t, a_t) = \sum_{s_{t+1}}{T(s_{t+1}|s_t, a_t)[R(s_t, s_{t+1})+\gamma V^\pi(s_{t+1})]}
\end{equation}

Unlike value function which specifies the goodness of a state, a Q-function specifies the goodness of action in a state.

\subsection{Category}
In general, RL can be divided into either model-free or model-based methods. Here, "model" is defined by the two quantity: transition probability function $T(s_{t+1} | s_t, a_t)$ and the reward function $R(s_t, s_{t+1})$. 

\subsubsection{Model-based RL} 
\begin{figure*}	
\centering
\begin{forest}
[Model-based RL
[\rotatebox{90}{Value Functions}
[\rotatebox{90}{Differential Dynamic Programming \cite{DDP_1}, \cite{DDP_2}}]
[\rotatebox{90}{Temporal Difference Learning \cite{TDP_1}}]
[\rotatebox{90}{Policy Iteration \cite{PI_1}}]
[\rotatebox{90}{Monte Carlo \cite{MTCL}}]
]
[\rotatebox{90}{Transition Models}
[\rotatebox{90}{Deterministic models}
[\rotatebox{90}{Decision trees \cite{TM_2}}]
[\rotatebox{90}{Linear regression \cite{TM_1}}]
]
[\rotatebox{90}{Stochastic models}
[\rotatebox{90}{Gaussian processes \cite{TM_3}, \cite{TM_4}, \cite{TM_5}}]
[\rotatebox{90}{Expectation-Maximization \cite{TM_6}} ]
[\rotatebox{90}{Dynamic Bayesian networks \cite{TM_2}}]
]
]
[\rotatebox{90}{Policy Search}
[\rotatebox{90}{Gradient-based \cite{PS_1}, \cite{PS_2}}
]
[\rotatebox{90}{Information theory \cite{TM_4}, \cite{PS_3}} 
]
[\rotatebox{90}{Sampling based \cite{RP_1}}
]
]
[\rotatebox{90}{Return Functions}
[\rotatebox{90}{Discounted returns functions \cite{RP_1}, \cite{RP_2}, \cite{RP_3}}
]
[\rotatebox{90}{Averaged returns functions \cite{RP_4}, \cite{RP_5}}
]
]
]
]
]
\end{forest}
\caption{Summarization of model-based RL approaches}
\label{fig:fig_summary}
\end{figure*}
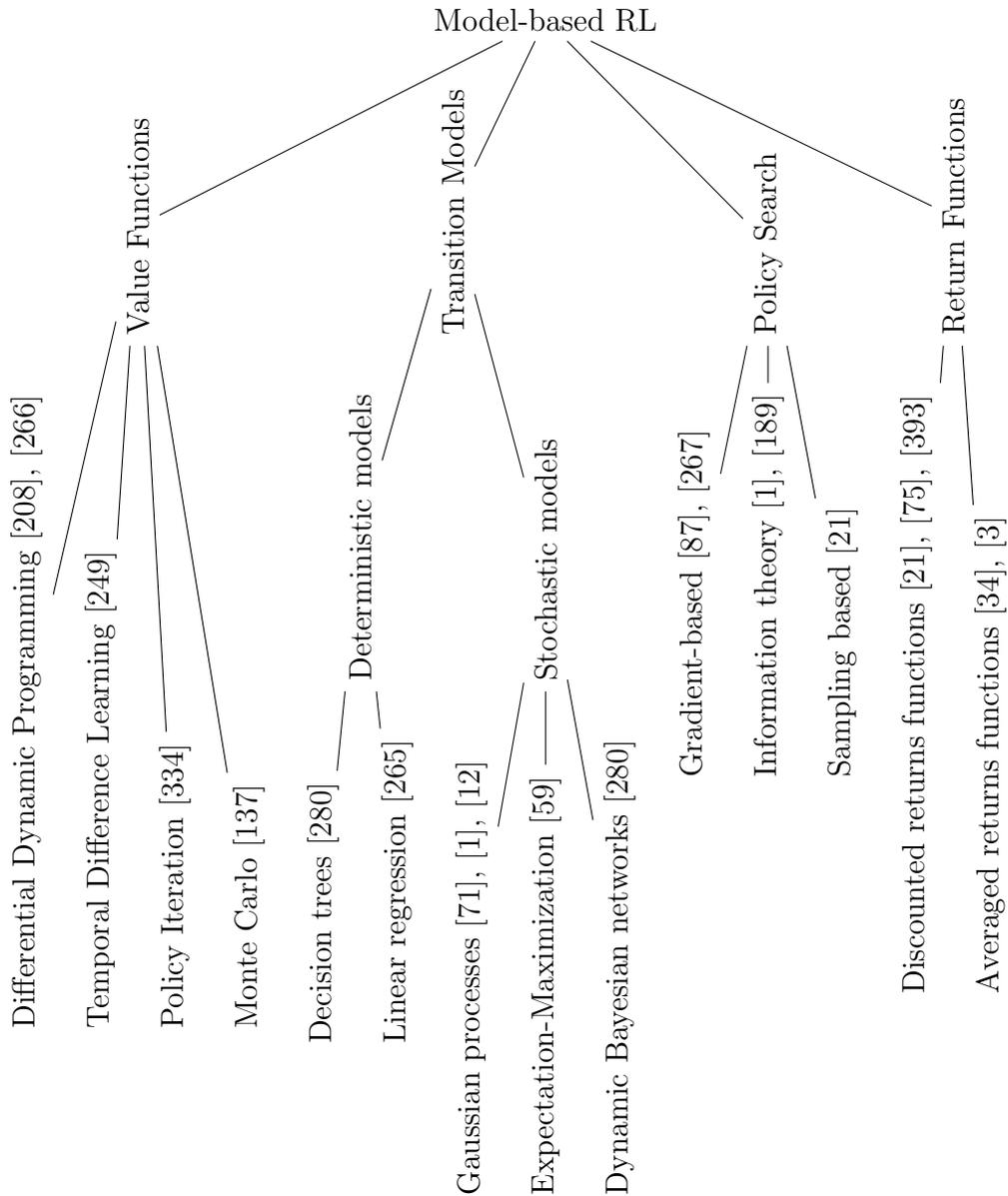

Model-based RL is an approach that uses a learnt model, i.e. $T(s_{t+1} | s_t, a_t) $ and reward function $R(s_t, s_{t+1})$ to predict the future action. There are four main model-based techniques as follows: 
\begin{itemize}
    \item \textbf{Value Function:} The objective of value function methods is to obtain the best policy by maximizing the value functions in each state. A value function of a RL problem can be defined as in Eq.\ref{eq:value} and the optimal state-value function is given in Eq.\ref{eq:bestvalue}  which are known as Bellman equations. Some common approaches in this group are Differential Dynamic Programming \cite{DDP_1}, \cite{DDP_2}, Temporal Difference Learning \cite{TDP_1}, Policy Iteration \cite{PI_1} and Monte Carlo \cite{MTCL}.

    \item \textbf{Transition Models:} Transition models decide how to map from a state s, taking action a to the next state (s') and it strongly affects the performance of model-based RL algorithms. Based on whether predicting the future state s' is based on the probability distribution of a random variable or not, there are two main approaches in this group: stochastic and deterministic. Some common methods for deterministic models are decision trees \cite{TM_2} and linear regression \cite{TM_1}. Some common methods for stochastic models are Gaussian processes \cite{TM_3}, \cite{TM_4}, \cite{TM_5}, Expectation-Maximization \cite{TM_6} and dynamic Bayesian networks \cite{TM_2}.
    
    \item \textbf{Policy Search:} Policy search approach directly searches for the optimal policy by modifying its parameters, whereas the value function methods indirectly find the actions that maximize the value function at each state. Some of the popular approaches in this group are: gradient-based \cite{PS_1}, \cite{PS_2}, information theory \cite{TM_4}, \cite{PS_3} and sampling based \cite{RP_1}. 
    
    \item \textbf{Return Functions:} Return functions decide how to aggregate rewards or punishments over an episode. They affect both the convergence and the feasibility of the model. There are two main approaches in this group: discounted returns functions \cite{RP_1}, \cite{RP_2}, \cite{RP_3} and averaged returns functions \cite{RP_4}, \cite{RP_5}. Between the two approaches, the former is the most popular which represents the uncertainty about future rewards. While small discount factors provide faster convergence, its solution may not be optimal. 
\end{itemize}
In practice, transition and reward functions are rarely known and hard to model. The comparative performance among all model-based techniques is reported in \cite{model-basedRL} with over 18 benchmarking environments including noisy environments. The Fig.\ref{fig:fig_summary} summarizes different model-based RL approaches.

\subsubsection{Model-free methods} Learning through the experience gained from interactions with the environment, i.e. model-free method tries to estimate the t. discrete problems transition probability function and the reward function from the experience to exploit them in acquisition of policy. Policy gradient and value-based algorithms are popularly used in model-free methods. 
\begin{itemize}
    \item \textbf{The policy gradient methods}: In this approach, RL task is considered as optimization with stochastic first-order optimization. Policy gradient methods directly optimize the discounted expected reward, i.e. $\mathcal{G}(\pi) \rightarrow \max_{\pi}$ to obtains the optimal policy $\pi^*$ without any additional information about MDP. To do so, approximate estimations of the gradient with respect to policy parameters are used. Take \cite{williams1992simple} as an example, policy gradient parameterizes the policy and updates parameters $\theta$,
    \begin{equation}
        \mathcal{G}^\theta(\pi) = \mathbb{E}_{{\mathcal{T}_\phi}}\sum_{t=0}{log( \pi_\theta(a_t|s_t))\gamma^{t}\mathcal{R}}
        \label{eq:policy_gradient1}
    \end{equation}
    where $\mathcal{R}$ is the total accumulated return and defined in Eq. \ref{eq:R}. Common used policies are Gibbs policies \cite{thesis_gibb}, \cite{gibb_2} and Gaussian policies \cite{gaussian_1}. Gibbs policies are used in discrete problems whereas Gaussian policies are used in continuous problems. 
    
    \item \textbf{Value-based methods}: In this approach, the optimal policy $\pi^*$ is implicitly conducted by gaining an approximation of optimal Q-function $Q^*(s, a)$. In value-based methods, agents update the value function to learn suitable policy while policy-based RL agents learn the policy directly. To do that,  Q-learning is a typical value-based method. The update rule of Q-learning with learning rate $\lambda$ is defined as:
    \begin{equation}
        Q(s_t, a_t) = Q(s_t, a_t) + \lambda \delta_t
    \end{equation}
    where $\delta_t = R(s_t, s_{t+1}) + \gamma \text{arg} \max_a{Q(s_{t+1}, a) - Q(s_{t}, a)}$ is the temporal difference (TD) error.
    
    Target at self-play Chess, \cite{value-basedRL} investigates inasmuch it is possible to leverage the qualitative feedback for learning an evaluation function for the game.  \cite{value-basedRL_2} provides the comparison of learning of linear evaluation functions between using preference learning and using least-squares temporal difference learning, from samples of game trajectories. The value-based methods depend on a specific, optimal policy, thus it is hard for transfer learning.
    
    \item \textbf{Actor-critic} is an improvement of policy gradient with an value-based critic $\Gamma$, thus, Eq.\ref{eq:policy_gradient1} is rewritten as:
    \begin{equation}
        \mathcal{G}^\theta(\pi) = \mathbb{E}_{{\mathcal{T}_\phi}}\sum_{t=0}{log( \pi_\theta(a_t|s_t))\gamma^{t}\Gamma_t }
        \label{eq:actor_crtic}
    \end{equation}
    The critic function $\Gamma$ can be defined as $Q^\pi(s_t, a_t)$ or $Q^\pi(s_t, a_t) - V^\pi_t $ or $R[s_{t-1}, s_{t}] + V^\pi_{t+1} - V^\pi_t$
\end{itemize}
    Actor-critic methods are combinations of actor-only methods and critic-only methods. Thus, actor-critic methods have been commonly used RL. Depend on reward setting, there are two groups of actor-critic methods, namely discounted return \cite{Actor-critic1}, \cite{bhatnagar2010actor} and average return \cite{Actor-critic3}, \cite{Actor-critic4}. 
 The comparison between model-based and model-free methods is given in Table \ref{tb:comparison}. 

\begin{table*}[]
\centering
\caption{Comparison between model-based RL and model-free RL}
\begin{tabular}{|l|l|l|}
\hline
\textbf{Factors}  & \textbf{Model-based RL}                                                       & \textbf{Model-free RL}                                                              \\ \hline
\begin{tabular}[c]{@{}l@{}}Number of iterations between \\ agent and environment\end{tabular} & Small  & Big  \\ \hline
Convergence & Fast  & Slow \\ \hline
Prior knowledge of transitions & Yes  & No \\ \hline
Flexibility                                        & \begin{tabular}[c]{@{}l@{}}Strongly depend on \\ a learnt model\end{tabular} & \begin{tabular}[c]{@{}l@{}}Adjust based \\ on trials and errors\end{tabular} \\ \hline
\end{tabular}
\label{tb:comparison}
\end{table*}



\section{Introduction to Deep Reinforcement Learning}

\label{sec:drl}
  DRL, which was proposed as a combination of RL and DL, has achieved rapid developments, thanks to the rich context representation of DL. Under DRL, the aforementioned value and policy can be expressed by neural networks which allow dealing with a continuous state or action that was hard for a table representation. Similar to RL, DRL can be categorized into model-based algorithms and model-free algorithms which will be introduced in this section. 

\subsection{Model-Free Algorithms}
There are two approaches, namely, Value-based DRL methods and Policy gradient DRL methods to implement model-free algorithms.

\subsubsection{Value-based DRL methods}
\begin{figure}[!h]
	\centering \includegraphics[width=0.8\columnwidth]{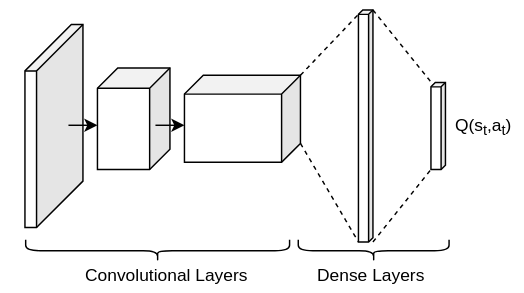}
		\caption{Network structure of Deep Q-Network (DQN), where Q-values Q(s,a) are generated for all actions for a given state. }
	\label{fig:dqn}
\end{figure}

\textbf{Deep Q-Learning Network (DQN):}
Deep Q-learning \cite{mnih2015human} (DQN) is the most famous DRL model which learns policies directly from high-dimensional inputs by CNNs. In DQN, input is raw pixels and output is a quality function to estimate future rewards as given in Fig.\ref{fig:dqn}. Take regression problem as an instance. Let $y$ denote the target of our regression task, the regression with input $(s, a)$, target $y(s, a)$ and the MSE loss function is as:
\begin{equation}
\begin{split}
    \mathcal{L^{DQN}} & = \mathcal{L}(y(s_t,a_t), Q^*(s_t, a_t, \theta_t)) \\
   & = ||y(s_t,a_t) - Q^*(s_t, a_t, \theta_t)||^2 \\
    y(s_t,a_t) & = R(s_t, s_{t+1}) + \gamma \max_{a_{t+1}}Q^*(s_{t_1}, a_{t+1}, \theta_t)
\end{split}
\label{eq:dqn}
\end{equation}
 Where $\theta$ is vector of parameters, $\theta \in \mathbb{R}^{|S||R|}$ and $s_{t+1}$ is a sample from $T(s_{t+1}| s_t, a_t)$ with input of $(s_t, a_t)$.

Minimizing the loss function yields a gradient descent step formula to update $\theta$ as follows:
\begin{equation}
    \theta_{t+1} = \theta_t -\alpha_t\frac{\mathcal{\partial L^{DQN}}}{\partial \theta}
\end{equation}

\noindent
\textbf{Double DQN:}
In DQN, the values of $Q^*$ in many domains were leading to overestimation because of $max$. In Eq.\ref{eq:dqn}, $y(s,a) = R(s, s') + \gamma \max_{a'}Q^*(s', a', \theta)$ shifts Q-value estimation towards either to the actions with high reward or to the actions with overestimating approximation error. Double DQN \cite{ddqn} is an improvement of DQN that combines double Q-learning \cite{hasselt2010double} with DQN and it aims at reducing observed overestimation with better performance. The idea of Double DQN is based on separating action selection and action evaluation using its own approximation of $Q^*$ as follows:
\begin{equation}
\begin{split}
\max_{a_{t+1}}Q^*(s_{t+1},a_{t+1};\theta) =  Q^*(s_{t+1}, \underset{a_{t+1}}{\arg\max}Q^*(s_{t+1},a_{t+1};\theta_1);\theta_2)
\end{split}
\end{equation}
Thus
\begin{equation}
y = R(s_t, s_{t+1}) + \gamma Q^*(s_{t+1}, \underset{a_{t+1}}{\arg\max}Q^*(s_{t+1},a_{t+1};\theta_1);\theta_2)
\end{equation}

The easiest and most expensive implementation of double DQN is to run two independent DQNs as follows:

\begin{equation}
\begin{split}
y_1 = R(s_t, s_{t+1}) + \\ \gamma Q^*_1(s_{t+1}, \underset{a_{t+1}}{\arg\max}Q^*_2(s_{t+1},a_{t+1};\theta_2);\theta_1) \\
y_2 = R(s_t, s_{t+1}) + \\ \gamma Q^*_2(s_{t+1}, \underset{a_{t+1}}{\arg\max}Q^*_1(s_{t+1},a_{t+1};\theta_1);\theta_2)
\end{split}
\end{equation}

\begin{figure}[!h]
	\centering \includegraphics[width=0.8\columnwidth]{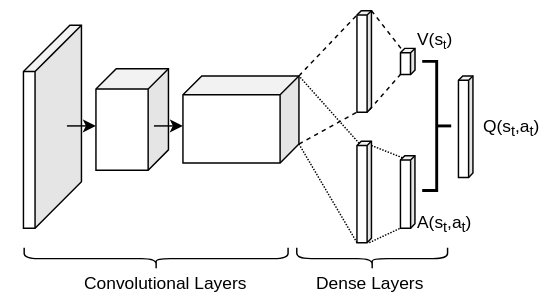}
		\caption{Network structure of Dueling DQN, where value function $V(s)$ and advantage function $A(s,a)$ are combined to predict Q-values $Q(s,a)$ for all actions for a given state. }
	\label{fig:ddqn}
\end{figure}

\noindent\textbf{Dueling DQN:}
In DQN, when the agent visits an unfavorable state, instead of lowering its value $V^*$, it remembers only low pay-off by updating $Q^*$. In order to address this limitation, Dueling DQN \cite{wang2015dueling} incorporates approximation of $V^*$ explicitly in a computational graph by introducing an advantage function as follows:
\begin{equation}
    A^{\pi}(s_t, a_t) = Q^{\pi}(s_t, a_t) - V^{\pi}(s_t)
\end{equation}
Therefore, we can reformulate Q-value: $Q^{*}(s, a) = A^{*}(s, a) + V^{*}(s)$. This implies that after DL the feature map is decomposed into two parts corresponding to $V^*(v)$ and $A^*(s, a)$ as illustrated in Fig.\ref{fig:ddqn}. This can be implemented by splitting the fully connected layers in the DQN architecture to compute the advantage and state value functions separately, then combining them back into a single Q-function. An interesting result has shown that Dueling DQN obtains better performance if it is formulated as:
\begin{equation}
    Q^{*}(s_t, a_t) =  V^{*}(s_t) + A^{*}(s_t, a_t) - \max_{a_{t+1}}A^{*}(s_t, a_{t+1})
\end{equation}
In practical implementation, averaging instead of maximum is used, i.e. 
\begin{equation}
\nonumber
    Q^{*}(s_t, a_t) =  V^{*}(s_t) + A^{*}(s_t, a_t) - \text{mean}_{a_{t+1}}A^{*}(s_t, a_{t+1})
\end{equation}
Furthermore, to address the limitation of memory and imperfect information at each decision point, Deep Recurrent Q-Network (DRQN) \cite{RDQN} employed RNNs into DQN by replacing the first fully-connected layer with an RNN. Multi-step DQN \cite{multi_step} is one of the most popular improvements of DQN by substituting one-step approximation with N-steps.

\noindent

\subsubsection{Policy gradient DRL methods}

\noindent
\textbf{Policy Gradient Theorem:}
Different from value-based DRL methods, policy gradient DRL optimizes the policy directly by optimizing the following objective function which is defined as a function of $\theta$. 
\begin{equation}
    \mathcal{G}(\theta) = \mathbb{E}_{\mathcal{T}\sim \pi_{\theta}} \sum_{t=1}{\gamma^{t-1}R(s_{t-1}, s_t)} \rightarrow \max_{\theta}
\label{eq:policy}
\end{equation}

\noindent For any MDP and differentiable policy $\pi_\theta$, the gradient of objective Eq.\ref{eq:policy} is defined by policy gradient theorem \cite{gradient_policy} as follows:
\begin{equation}
    \bigtriangledown_\theta \mathcal{G}(\theta) = \mathbb{E}_{\mathcal{T}\sim\pi_{\theta}} \sum_{t=0}{\gamma^{t}Q^\pi(s_t,a_t)\bigtriangledown_\theta \text{log} \pi_{\theta}(a_t| s_t)} 
\label{eq:policy_gradient}
\end{equation}
\noindent
\textbf{REINFORCE:}
REINFORCE was introduced by \cite{williams1992simple} to approximately calculate the gradient in  Eq.\ref{eq:policy_gradient} by using Monte-Carlo estimation. In REINFORCE approximate estimator, Eq.\ref{eq:policy_gradient} is reformulated as:
\begin{equation}
    \bigtriangledown_\theta \mathcal{G}(\theta) \approx \sum_{\mathcal{T}}^N \sum_{t=0}{\gamma^{t}\bigtriangledown_\theta \text{log} \pi_{\theta}(a_t| s_t)(\sum_{t'=t}{\gamma^{t'-t}R(s_{t'}, s_{t'+1})})} 
\label{reinforce}
\end{equation}
where $\mathcal{T}$ is trajectory distribution and defined in Eq.\ref{eq:T0}. Theoretically, REINFORCE can be straightforwardly applied into any parametric $\pi_{theta}(a|s)$. However, it is impractical to use because of its time-consuming nature for convergence and local optimums problem. Based on the observation that the convergence rate of stochastic gradient descent directly depends on the variance of gradient estimation, the variance reduction technique was proposed to address naive REINFORCE's limitations by adding a term that reduces the variance without affecting the expectation.



\subsubsection{Actor-Critic DRL algorithm}
\begin{figure}[!h]
	\centering \includegraphics[width=0.5\columnwidth]{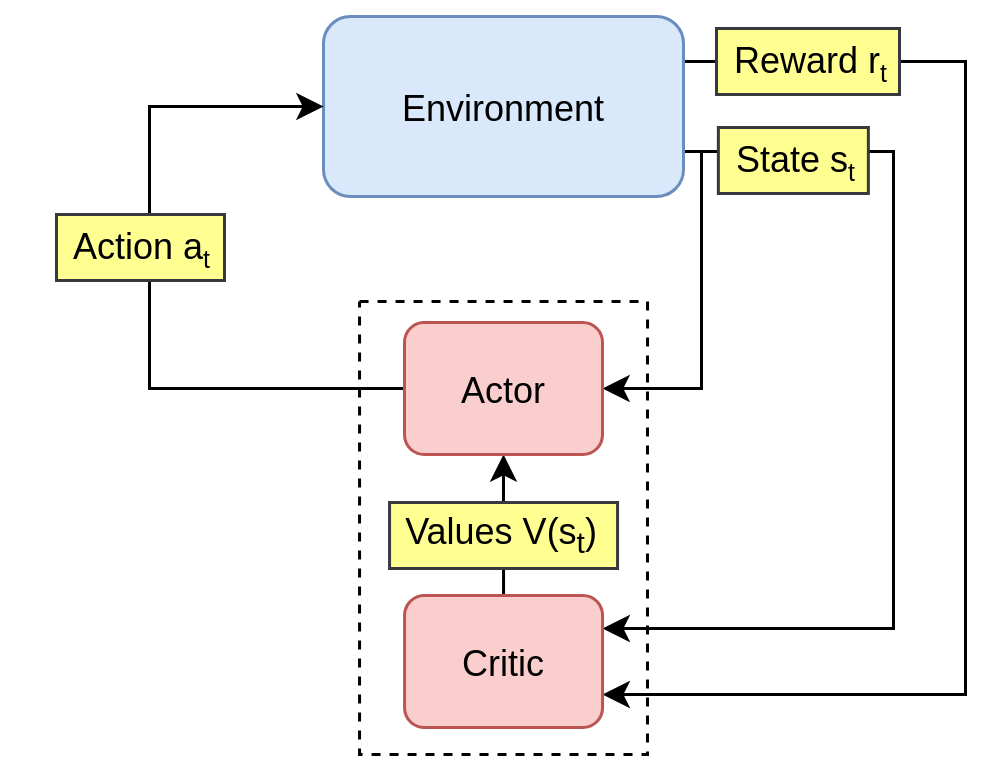}
		\caption{Flowchart showing the structure of actor critic algorithm.}
	\label{fig:a3c}
\end{figure}

Both value-based and policy gradient algorithms have their own pros and cons, i.e. policy gradient methods are better for continuous and stochastic environments, and have a faster convergence whereas, value-based methods are more sample efficient and steady. Lately, actor-critic \cite{konda2000actor} \cite{mnih2016asynchronous} was born to take advantage from both value-based and policy gradient while limiting their drawbacks. Actor-critic architecture computes the policy gradient using a value-based critic function to estimate expected future reward. The principal idea of actor-critic is to divide the model into two parts: (i) computing an action based on a state and (ii) producing the Q values of the action. As given in Fig.\ref{fig:a3c}, the actor takes as input the state $s_t$ and outputs the best action $a_t$. It essentially controls how the agent behaves by learning the optimal policy (policy-based). The critic, on the other hand, evaluates the action by computing the value function (value-based). The most basic actor-critic method (beyond the tabular case) is naive policy gradients (REINFORCE). The relationship between actor-critic is similar to kid-mom. The kid (actor) explores the environment around him/her with new actions i.e. tough fire, hit a wall, climb a tree, etc while the mom (critic) watches the kid and criticizes/compliments him/her. The kid then adjusts his/her behavior based on what his/her mom told. When the kids get older, he/she can realize which action is bad/good.

\begin{figure}[!h]
	\centering \includegraphics[width=0.9\columnwidth]{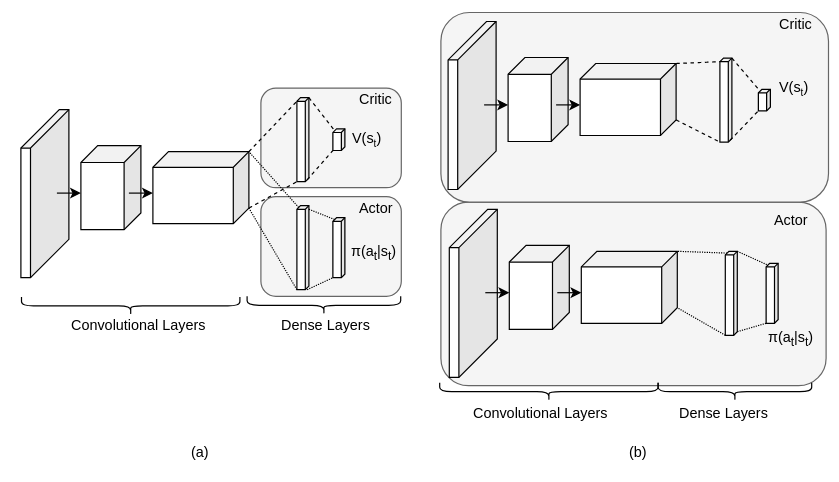}
		\caption{An illustration of Actor-Critic algorithm in two cases: sharing parameters (a) and not sharing parameters (b).}
	\label{fig:a3cn}
\end{figure}

\noindent
\textbf{Advantage Actor-Critic (A2C)}
Advantage Actor-Critic (A2C) \cite{a3c}  consist of two neural networks i.e. actor network $\pi_\theta(a_t|s_t)$ representing for policy and critic network $V^\pi_\omega$ with parameters $\omega$ approximately estimating actor’s performance. 
In order to determine how much better, it is to take a specific action compared to the average, an advantage value is defined as: 
\begin{equation}
    A^\pi(s_t, a_t) = Q^\pi(s_t, a_t) - V^\pi(s_t)
\end{equation}
Instead of constructing two neural networks for both the Q value and the V value, using the Bellman optimization equation, we can rewrite the advantage function as:
\begin{equation}
    A^\pi(s_t, a_t) = R(s_t, s_{t+1})+  \gamma V^\pi_\omega(s_{t+1}) - V^\pi_\omega(s_t)
\label{eq:A2C_step1}
\end{equation}
For given policy $\pi$, its value function can be obtained using point iteration for solving:
\begin{equation}
    V^\pi(s_t) = \mathbb{E}_{a_t\sim \pi( a_t| s_t)}\mathbb{E}_{s_{t+1}\sim T(s_{t+1}|a_t,s_t)}(R(s_t, s_{t+1}) + \gamma V^\pi(s_{t+1}))
    \label{eq:A2C_V}
\end{equation}
Similar to DQN, on each update a target is computed using current approximation: 
\begin{equation}
y = R(s_t, s_{t+1}) + \gamma V^\pi_\omega(s_{t+1})
\label{eq:A2C_step2}
\end{equation}
 At time step t, the A2C algorithm can be implemented as following steps: 
 \begin{itemize}
     \item Step 1: Compute advantage function using Eq.\ref{eq:A2C_step1}.
     \item Step 2: Compute target using Eq.\ref{eq:A2C_step2}.
     \item Step 3: Compute critic loss with MSE loss: $\mathcal{L} = \frac{1}{B}\sum_T||y - V^\pi(s_t))||^2$, where $B$ is batch size and $V^\pi(s_t)$ is defined in Eq.\ref{eq:A2C_V}.
     \item Step 4: Compute critic gradient: $\bigtriangledown^{critic} = \frac{\partial \mathcal{L}}{\partial \omega}$.
     \item Step 5: Compute actor gradient: $\bigtriangledown^{actor} = \frac{1}{B}\sum_T{\bigtriangledown_\theta \text{log} \pi(a_t|s_t)A^\pi(s_t, a_t)}$
 \end{itemize}

\noindent
\textbf{Asynchronous Advantage Actor Critic (A3C)}
Besides A2C, there is another strategy to implement an Actor-Critic agent. Asynchronous Advantage Actor-Critic (A3C) \cite{a3c} approach does not use experience replay because this requires a lot of memory. Instead, A3C asynchronously executes different agents in parallel on multiple instances of the environment. Each worker (copy of the network) will update the global network asynchronously. Because of the asynchronous nature of A3C, some workers (copy of the agents) will work with older values of the parameters. Thus the aggregating update will not be optimal. On the other hand, A2C synchronously updates the global network. A2C waits until all workers finished their training and calculated their gradients to average them, to update the global network. In order to update the entire network, A2C waits for each actor to finish their segment of experience before updating the global parameters. As a consequence, the training will be more cohesive and faster. Different from A3C, each worker in A2C has the same set of weights since and A2C updates all their workers at the same time. In short, A2C is an alternative to the synchronous version of the A3C. In A2C, it waits for each actor to finish its segment of experience before updating, averaging over all of the actors. In a practical experiment, this implementation is more effectively uses GPUs due to larger batch sizes. The structure of an actor-critic algorithm can be divided into two types depending on parameter sharing as illustrated in Fig.\ref{fig:a3cn}. 

In order to overcome the limitation of speed, GA3C \cite{ga3c} was proposed and it achieved a significant speedup compared to the original CPU implementation. To more effectively train A3C, \cite{ftfe} proposed FFE which forces random exploration at the right time during a training episode, that can lead to improved training performance. 


\subsection{Model-Based Algorithms}
We have discussed so far model-free methods including the value-based approach and policy gradient approach. In this section, we focus on the model-based approach, that deals with the dynamics of the environment by learning a transition model that allows for simulation of the environment without interacting with the environment directly. In contrast to model-free approaches, model-based approaches are learned from experience by a function approximation. Theoretically, no specific prior knowledge is required in model-based RL/DRL but incorporating prior knowledge can help faster convergence and better-trained model, speed up training time as well as the number of training samples. While using raw data with pixel, it is difficult for model-based RL to work on high dimensional and dynamic environments. This is addressed in DRL by embedding the high-dimensional observations into a lower-dimensional space using autoencoders \cite{model_based_DRL}. Many DRL approaches have been based on scaling up prior work in RL to high-dimensional problems.  A good overview of model-based RL for high-dimensional problems can be found in \cite{plaat2020deep} which partition model-based DRL into three categories: explicit planning on given transitions, explicit planning on learned transitions, and end-to-end learning of both planning and transitions. In general, DRL targets training DNNs to approximate the optimal policy $\pi^*$ together with optimal value functions $V^*$ and $Q^*$. In the following, we will cover the most common model-based DRL approaches including value function and policy search methods. 
\subsubsection{Value function}
We start this category with DQN \cite{mnih2015human} which has been successfully applied to classic Atari and illustrated in Fig.\ref{fig:dqn}. DQN uses CNNs to deal with high dimensional state space like pixels, to approximate the Q-value function. 

\noindent
\textbf{Monte Carlo tree search (MCTS)} MCTS \cite{mtcs} is one of the most popular methods to look-ahead search and it is combined with a DNN-based transition model to build a model-based DRL in \cite{model-based-value_DRL0}. In this work, the learned transition model predicts the next frame and the rewards one step ahead using the input of the last four frames of the agent’s first-person-view image and the current action. This model is then used by the Monte Carlo tree search algorithm to plan the best sequence of actions for the agent to perform. 

\noindent 
\textbf{Value-Targeted Regression (UCRL-VTR)} Alex, et al. proposed model-based DRL for regret minimization \cite{model-based-value_DRL}. In their work, a set of models, that are ‘consistent’ with the data collected, is constructed at each episode. The consistency is defined as the total squared error, whereas the value function is determined by solving the optimistic planning problem with the constructed set of models

\subsubsection{Policy search}
Policy search methods aim to directly find policies by means of gradient-free or gradient-based methods.

\noindent
\textbf{Model-Ensemble Trust-Region Policy Optimization (ME-TRPO)} ME-TRPO \cite{model-based-value_DRL1} is mainly based on Trust Region Policy Optimization (TRPO) \cite{trpo} which imposes a trust region constraint on the policy to further stabilize learning. 
 
\noindent
\textbf{Model-Based Meta-Policy-Optimization (MB-MPO)} MB-MPO \cite{model-based-policy_DRL} addresses the performance limitation of model-based DRL compared against model-free DRL when learning dynamics models. MB-MPO learns an ensemble of dynamics models, a policy that can quickly adapt to any model in the ensemble with one policy gradient step. As a result, the learned policy exhibits less model bias without the need to behave conservatively.


A summary of both model-based and model-free DRL algorithms is given in Table \ref{tab:sum_drl}. In this Table, we also categorized DRL techniques into either on-policy or off-policy. In on-policy RL, it allows the use of older samples (collected using the older policies) in the calculation. The policy $\pi^k$ is updated with data collected by $\pi^k$ itself. In off-policy RL, the data is assumed to be composed of different policies $\pi^0, \pi^0, ..., \pi^k$. Each policy has its own data collection, then the data collected from  $\pi^0$,  $\pi^1$, ...,  $\pi^k$ is used to train $\pi^{k+1}$.

\begin{table*}
\caption{Summary of model-based and model-free DRL algorithms consisting of value-based and policy gradient methods.}
\begin{tabular}{l|l|l}
    DRL Algorithms & Description & Category \\\hline
    DQN \cite{mnih2015human}& Deep Q Network & \shortstack{Value-based\\ Off-policy}\\
    Double DQN \cite{ddqn} & Double Deep Q Network & \shortstack{Value-based\\ Off-policy} \\ 
    Dueling DQN \cite{wang2015dueling} & Dueling Deep Q Network & \shortstack{Value-based\\ Off-policy} \\
    MCTS \cite{model-based-value_DRL0} & Monte Carlo tree search & \shortstack{Value-based\\ On-policy} \\
    UCRL-VTR\cite{model-based-value_DRL} & optimistic planning problem & \shortstack{Value-based\\ Off-policy}\\
    DDPG \cite{ddpg} & DQN with Deterministic Policy Gradient & \shortstack{Policy gradient \\ Off-policy} \\
    TRPO \cite{trpo} & Trust Region Policy Optimization & \shortstack{Policy gradient\\On-policy} \\
    PPO \cite{ppo} & Proximal Policy Optimization & \shortstack{Policy gradient\\ On-policy} \\
    ME-TRPO \cite{model-based-value_DRL1} & Model-Ensemble Trust-Region Policy Optimization & \shortstack{Policy gradient \\ On-policy} \\
    MB-MPO \cite{model-based-policy_DRL} & Model-Based Meta- Policy-Optimization & \shortstack{Policy gradient\\ On-policy}\\
    A3C \cite{a3c} & Asynchronous Advantage Actor Critic & \shortstack{Actor Critic\\ On-Policy} \\
    A2C \cite {a3c} & Advantage Actor Critic & \shortstack{Actor Critic\\ On-Policy} \\
\end{tabular}
\label{tab:sum_drl}
\end{table*}

\subsection{Good practices}
Inspired by Deep Q-learning \cite{mnih2015human}, we discuss some useful techniques that are used during training an agent in DRL framework in practices.

\noindent
\textbf{Experience replay}
Experience replay \cite{zha2019experience} is a useful part of off-policy learning and is often used while training an agent in RL framework. By getting rid of as much information as possible from past experiences, it removes the correlations in training data and reduces the oscillation of the learning procedure. As a result, it enables agents to remember and re-use past experiences sometimes in many weights updates which increases data efficiency.

\noindent
\textbf{Minibatch learning}
Minibatch learning is a common technique that is used together with experience replay. Minibatch allows learning more than one training sample at each step, thus, it makes the learning process robust to outliers and noise.

\noindent
\textbf{Target Q-network freezing}
As described in \cite{mnih2015human}, two networks are used for the training process. In target Q-network freezing: one network interacts with the environment and another network plays the role of a target network. The first network is used to generate target Q-values that are used to calculate losses. The weights of the second network i.e. target network are fixed and slowly updated to the first network \cite{lillicrap2015continuous}.

\noindent
\textbf{Reward clipping}
A reward is the scalar number provided by the environment and it aims at optimizing the network. To keep the rewards in a reasonable scale and to ensure proper learning, they are clipped to a specific range (-1 ,1). Here 1 refers to as positive reinforcement or reward and -1 is referred to as negative reinforcement or punishment.

\noindent
\textbf{Model-based v.s. model-free approach}
Whether the model-free or model-based approaches is chosen mainly depends on the model architecture i.e. policy and value function.

\section{DRL in Landmark Detection}
\label{sec:landmarkdetection}
Autonomous landmark detection has gained more and more attention in the past few years. One of the main reasons for this increased inclination is the rise of automation for evaluating data. The motivation behind using an algorithm for landmarking instead of a person is that manual annotation is a time-consuming tedious task and is prone to errors. Many efforts have been made for the automation of this task. Most of the works that were published for this task using a machine learning algorithm to solve the problem. \cite{criminisi2010regression} proposed a regression forest-based method for detecting landmark in a full-body CT scan. Although the method was fast it was less accurate when dealing with large organs. \cite{gauriau2014multi} extended the work of \cite{criminisi2010regression} by adding statistical shape priors that were derived from segmentation masks with cascade regression.

In order to address the limitations of previous works on anatomy detection, \cite{ghesu2017multi} reformulated the detection problem as a behavior learning task for an artificial agent using MDP. By using the capabilities of DRL and scale-space theory \cite{lindeberg2013scale}, the optimal search strategies for finding anatomical structures are learned based on the image information at multiple scales. In their approach, the search starts at the coarsest scale level for capturing global context and continues to finer scales for capturing more local information. In their RL configuration, the state of the agent at time $t$, $s_t = I(\vec{p}_t)$ is defined as an axis-aligned box of image intensities extracted from the image $I$ and centered at the voxel-position $\vec{p}_t$ in image space. An action $a_t$ allows the agent to move from any voxel position $\vec{p}_t$ to an adjacent voxel position $\vec{p}_{t+1}$. The reward function represents distance-based feedback, which is positive if the agent gets closer to the target structure and negative otherwise. In this work, a CNN is used to extract deep semantic features. The search starts with the coarsest scale level $M-1$, the algorithm tries to maximize the reward which is the change in distance between ground truth and predicted landmark location before and after the action of moving the scale window across the image. Upon convergence, the scale level is changed to $M-2$ and the search continued from the convergence point at scale level $M-1$. The process is repeated on the following scales until convergence on the finest scale. The authors performed experiments on 3D CT scans and obtained an average accuracy increase of 20-30$\%$ and lower distance error than the other techniques such as SADNN \cite{ghesu2016marginal} and 3D-DL \cite{zheng20153d}

Focus on anatomical landmark localization in 3D fetal US images, \cite{alansary2019evaluating} proposed and demonstrated use cases of several different Deep Q-Network RL models to train agents that can precisely localize target landmarks in medical scans. In their work, they formulate the landmark detection problem as an MDP of a goal-oriented agent, where an artificial agent is learned to make a sequence of decisions towards the target point of interest. At each time step, the agent should decide which direction it has to move to find the target landmark. These sequential actions form a learned policy forming a path between the starting point and the target landmark. This sequential decision-making process is approximated under RL. In this RL configuration, the environment is defined as a 3D input image, action $A$ is a set of six actions {$a_x+, a_x-, a_y+, a_y-, a_z+, a_z-$} corresponding to three directions, the state $s$ is defined as a 3D region of interest (ROI) centered around the target landmark and the reward is chosen as the difference between the two Euclidean distances: the previous step and current step. This reward signifies whether the agent is moving closer to or further away from the desired target location. In this work, they also proposed a novel fixed- and multi-scale optimal path search strategy with hierarchical action steps for agent-based landmark localization frameworks. 

Whereas pure policy or value-based methods have been widely used to solve RL-based localization problems, \cite{al2019partial} adopts an actor-critic \cite{mnih2016asynchronous} based direct policy search method framed in a temporal difference learning approach. In their work, the state is defined as a function of the agent-position which allows the agent at any position to observe an $m \times m\times 3$ block of surrounding voxels. Similar to other previous work, the action space is {$a_x+, a_x-, a_y+, a_y-, a_z+, a_z-$}. The reward is chosen as a simple binary reward function, where a positive reward is given if an action leads the agent closer to the target landmark, and a negative reward is given otherwise. Far apart from the previous work, their approach proposes a non-linear policy function approximator represented by an MLP whereas the value function approximator is presented by another MLP stacked on top of the same CNN from the policy net. Both policy (actor) and value (critic) networks are updated by actor-critic learning. To improve the learning, they introduce a partial policy-based RL to enable solving the large problem of localization by learning the optimal policy on smaller partial domains. The objective of the partial policy is to obtain multiple simple policies on the projections of the actual action space, where the projected policies can reconstruct the policy on the original action space. 

Based on the hypothesis that the position of all anatomical landmarks is interdependent and non-random within the human anatomy and this is necessary as the localization of different landmarks requires learning partly heterogeneous policies, \cite{vlontzos2019multiple} concluded that one landmark can help to deduce the location of others. For collective gain, the agents share their accumulated knowledge during training. In their approach, the state is defined as RoI centered around the location of the agent. The reward function is defined as the relative improvement in Euclidean distance between their location at time $t$ and the target landmark location. Each agent is considered as Partially Observable Markov Decision Process (POMDP) \cite{girard2015concurrent} and calculates its individual reward as their policies are disjoint. In order to reduce the computational load in locating multiple landmarks and increase accuracy through anatomical interdependence, they propose a collaborative multi-agent landmark detection framework (Collab-DQN) where DQN is built upon a CNN. The backbone CNN is shared across all agents while the policy-making fully connected layers are separate for each agent.

\begin{spacing}{1.1}
\footnotesize
\begin{longtable}{|p{0.09\linewidth}|p{0.04\linewidth}|p{0.08\linewidth}|p{0.09\linewidth}|p{0.19\linewidth}|p{0.20\linewidth}|p{0.15\linewidth}|}

\caption{Comparing various DRL-based landmark detection methods. The first group on Single Landmark Detection (SLD) and the second group for Multiple Landmark Detection (MLD)} 

\label{tab:landmark}
\\
\hline
 Approaches &
  Year &
 Training Technique &
  Actions &
  Remarks &
  Performance &
  Datasets and source code  \\ \hline \hline
SLD \cite{ghesu2017multi}&
  2017 &
  DQN &
 6 action: 2 per axis &
   State: an axis-aligned box centered at the voxel-position. Action: move from  $\vec{p}_t$ to $\vec{p}_{t+1}$. 
  Reward: distance-based feedback &
  Average accuracy increase 20-30\%. Lower distance error than other techniques such as SADNN \cite{ghesu2016marginal} and 3D-DL \cite{zheng20153d} &
  3D CT Scan \\ \hline
  
SLD \cite{alansary2019evaluating} &
  2019 &
  DQN, DDQN, Duel DQN and Duel DDQN 
  &
  6 action: 2 per axis&
  Environment: 3D input image. 
  State: 3D RoI centered around the target landmark. Reward: Euclidean distance between predicted points and groundtruth points. 
  &
  
  Duel DQN performs the best
  on Right Cerebellum (FS),
  Left Cerebellum (FS, MS)
  Duel DDQN is the best 
  on Right Cerebellum (MS)
  DQN performs the best 
  on Cavum Septum Pellucidum(FS, MS)
  &
  Fetal head, ultrasound scans \cite{li2018fast}. \newline \href{https://github.com/amiralansary/rl-medical}{Available code}
  \\
  \hline
SLD \cite{al2019partial} &
  2019 &
  Actor- Critic -based Partial -Policy RL &
  6 action: 2 per axis &
  State: a function of the agent-position.
  Reward:  binary reward function. policy function: MLP. value function: MLP &
  Faster and better convergence, 
  outperforms than other conventional actor-critic and Q-learning &
  CT volumes: Aortic valve. CT volumes: LAA seed-point. MR images: Vertebra centers \cite{cai2015multi}.\\ \hline \hline

MLD \cite{vlontzos2019multiple} &
  2019 &
  Collab DQN &
  6 action: 2 per axis &
  State: RoI centred around the agent.
  Reward: relative improvement in Euclidean distance. Each Agent is a POMDP has its own reward. Collab-DQN: reduce the computational load &
  Colab DQN got better results than supervised CNN and DQN &
  Brain MRI landmark \cite{jack2008alzheimer}, Cardiac MRI landmark \cite{de2014population}, Fetal brain landmark \cite{alansary2019evaluating}. \newline \href{https://github.com/thanosvlo/MARL-for-Anatomical-Landmark-Detection}{Available code} \\ \hline

 MLD \cite{jain2020robust} &
  2020 &
  DQN &
  6 action  2 per axis &
  State: 3D image patch. 
  Reward: Euclidean distance and $\in [-1, 1]$. Backbone CNN is share among agents Each agent has it own Fully connected layer &
  Detection error increased as the degree of missing information increased Performance is affected by the choice of landmarks
  &
  3D Head MR images \\  \hline


\end{longtable}

\end{spacing}

Different from the previous works on RL-based landmark detection, which detect a single landmark,\cite{jain2020robust} proposed a multiple landmark detection approach to better time-efficient and more robust to missing data. In their approach, each landmark is guided by one agent. The MDP is models as follows: The state is defined as a 3D image patch. The reward, clipped in [-1, +1], is defined as the difference in the Euclidean distance between the landmark predicted in the previous time step and the target, and in the landmark predicted in the current time step and the target. The action space is defined as in other previous works i.e. there are 6 actions {$a_x+, a_x-, a_y+, a_y-, a_z+, a_z-$} in the action space. To enable the agents to share the information learned by detecting one landmark for use in detecting other landmarks, hard parameter sharing from multi-task learning is used. In this work, the backbone network is shared among agents and each agent has its own fully connected layer. 

\begin{figure}[!h]
	\centering \includegraphics[width=1.0\columnwidth]{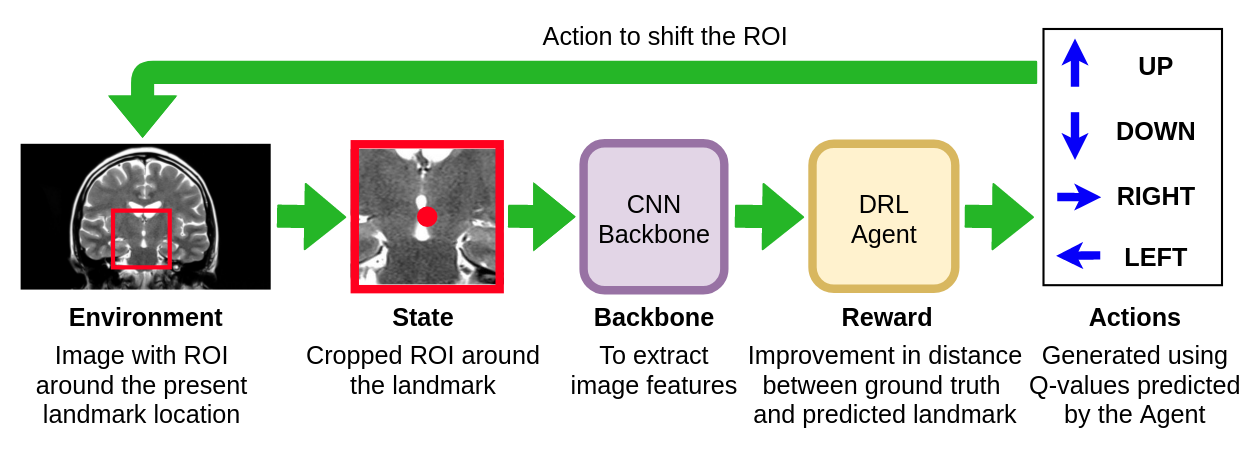}
		\caption{DRL implementation for landmark detection, The red point corresponds to the current landmark location and Red box is the Region of Interest (ROI) centered around the landmark, the actions of DRL agent shift the ROI across the image to maximize the reward corresponding to the improvement in distance between the ground truth and predicted landmark location. }
	\label{fig:land}
\end{figure}

Table \ref{tab:landmark} summarizes and compares all approaches for DRL in landmark detection, and a basic implementation of landmark detection using DRL has been shown in Fig. \ref{fig:land}. The figure illustrates a general implementation of landmark detection with the help of DRL, where the state is the Region of interest (ROI) around the current landmark location cropped from the image, The actions performed by the DRL agent are responsible for shifting the ROI across the image forming a new state and the reward corresponds to the improvement in euclidean distance between ground truth and predicted landmark location with iterations as used by \cite{ghesu2017multi},\cite{al2019partial},\cite{alansary2019evaluating},\cite{vlontzos2019multiple},\cite{jain2020robust}.

\section{DRL in Object Detection }
\label{sec:objectdetection}

Object detection is a task that requires the algorithm to find bounding boxes for all objects in a given image. Many attempts have been made towards object detection. A method for bounding box prediction for object detection was proposed by  \cite{girshick2014rich}, in which the task was performed by extracting region proposals from an image and then feeding each of them to a CNN to classify each region. An improvement to this technique was proposed by \cite{girshick2015fast}, where they used the feature from the CNN to propose region proposals instead of the image itself, this resulted in fast detection. Further improvement was proposed by \cite{ren2015faster}, where the authors proposed using a region proposal network (RPN) to identify the region of interest, resulting in much faster detection. Other attempts including focal loss \cite{lin2017focal} and Fast YOLO \cite{shafiee2017fast} have been proposed to address the imbalanced data problem in object detection with focal loss \cite{lin2017focal}, and perform object detection in video on embedded devices in a real-time manner \cite{shafiee2017fast}.

Considering MDP as the framework for solving the problem, \cite{caicedo2015active} used DRL for active object localization. The authors considered 8 different actions (up, down, left, right, bigger, smaller, fatter, taller) to improve the fit of the bounding box around the object and additional action to trigger the goal state. They used a tuple of feature vector and history of actions for state and change in IOU across actions as a reward.


An improvement to \cite{caicedo2015active} was proposed by \cite{bellver2016hierarchical}, where the authors used a hierarchical approach for object detection by treating the problem of object detection as an MDP. In their method, the agent was responsible to find a region of interest in the image and then reducing the region of interest to find smaller regions from the previously selected region and hence forming a hierarchy. For the reward function, they used the change in Intersection over union (IOU) across the actions and used DQN as the agent. As described in their paper, two networks namely, Image-zooms and Pool45-crops with VGG-16 \cite{simonyan2014very} backbone were used to extract the feature information that formed the state for DQN along with a memory vector of the last four actions.

Using a sequential search strategy, \cite{mathe2016reinforcement} proposed a method for object detection using DRL. The authors trained the model with a set of image regions where at each time step the agent returned fixate actions that specified a location in image for actor to explore next and the terminal state was specified by $done$ action. The state consisted of a tuple three elements: the observed region history $H_{t}$, selected evidence region history $E_{t}$ and fixate history $F_{t}$. The $fixate$ action was also a tuple of three elements: $fixate$ action, index of evidence region $e_{t}$ and image coordinate of next fixate $z_{t}$. The $done$ action consisted of: $done$ action, index of region representing the detected output $b_{t}$ and the detection confidence $c_{t}$. The authors defined the reward function that was sensitive to the detection location, the confidence at the final state and incurs a penalty for each region evaluation.

To map the inter-dependencies among the different objects, \cite{jie2016tree} proposed a tree-structured RL agent (Tree-RL) for object localization by considering the problem as an MDP. The authors in their implementation considered actions of two types: translation and scaling, where the scaling consisted of five actions whereas translation consisted of eight actions. In the specified work, the authors used the state as a concatenation of the feature vector of the current window, feature vector of the whole image, and history of taken actions. The feature vector were extracted from an ImageNet \cite{deng2009imagenet} \cite{russakovsky2015imagenet} trained VGG-16 \cite{simonyan2014very} model and for reward the change in IOU across an action was used. Tree-RL utilized a top-down tress search starting from the whole image where each window recursively takes the best action from each action group which further gives two new windows. This process is repeated recursively to find the object.

The task of breast lesion detection is a challenging yet very important task in the medical imaging field. A DRL method for active lesion detection in the breast was proposed by \cite{maicas2017deep}, where the authors formulated the problem as an MDP. In their formulation, a total of nine actions consisting of 6 translation actions, 2 scaling actions, and 1 trigger action were used. In the specified work, the change in dice coefficient across an action was used as the reward for scaling and translation actions, and for trigger action, the reward was $+\eta$ for dice coefficient greater than $r_{w}$ and $-\eta$ otherwise, where $\eta$ and $r_{w}$ were the hyperparameters chosen by the authors. For network structure, ResNet \cite{he2016deep} was used as the backbone and DQN as the agent.

Different from the previous methods, \cite{wang2018multitask} proposed a method for multitask learning using DRL for object localization. The authors considered the problem as an MDP where the agent was responsible to perform a series of transformations on the bounding box using a series of actions. Utilizing an RL framework the states consisted of feature vector and historical actions concatenated together, and a total of 8 actions for Bounding box transformation (left, right, up, down, bigger, smaller, fatter, and taller) were used. For reward the authors used the change in IOU between actions, the reward being 0 for an increase in IOU and -1 otherwise. For terminal action, however, the reward was 8 for IOU greater than 0.5 and -8 otherwise. The authors in the paper used DQN with multitask learning for localization and divided terminal action and 8 transformation actions into two networks and trained them together.

An improvement for the Region proposal networks that greedily select the ROIs was proposed by   \cite{pirinen2018deep}, where they used RL for the task. The authors in this paper used a two-stage detector similar to Fast and Faster R-CNN But used RL for the decision-making Process. For the reward, they used the normalized change in Intersection over Union (IOU).

Instead of learning a policy from a large set of data, \cite{aylebar} proposed a method for bounding box refinement (BAR) using RL. In the paper, once the authors have an inaccurate bounding box that is predicted by some algorithm they use the BAR algorithm to predict a series of actions for refinement of a bounding box. They considered a total of 8 actions (up, down, left, right, wider, taller, fatter, thinner) for bounding box transformation and considered the problem as a sequential decision-making problem (SDMP). They proposed an offline method called BAR-DRL and an online method called BAR-CB where training is done on every image. In BAR-DRL the authors trained a DQN over the states which consisted of features extracted from ResNet50 \cite{he2016deep} \cite{szegedy2017inception} pretrained on ImageNet \cite{deng2009imagenet} \cite{russakovsky2015imagenet} and a history vector of 10 actions. The Reward for BAR-DRL was 1 if the IOU increase after action and -3 otherwise. For BAR-CB they adapted the LinUCB \cite{li2010contextual} algorithm for an episodic scenario and considered The Histogram of Oriented Gradients (HOG) for the state to capture the outline and edges of the object of interest. The actions in the online method (BAR-CB) were the same as the offline method and the reward was 1 for increasing IOU and 0 otherwise. For both the implementations, the authors considered $\beta$ as terminal IOU. 

An improvement to sequential search strategy by \cite{mathe2016reinforcement} was proposed by \cite{uzkent2020efficient}, where they used a framework consisting of two modules, Coarse and fine level search. According to the authors, this method is efficient for object detection in large images (dimensions larger than 3000 pixels). The authors first performed a course level search on a large image to find a set of patches that are used by fine level search to find sub-patches. Both fine and coarse levels were conducted using a two-step episodic MDP, where The policy network was responsible for returning the probability distribution of all actions. In the paper, the authors considered the actions to be the binary action array (0,1) where 1 means that the agent would consider acquiring sub-patches for that particular patch. The authors in their implementation considered a number of patches and sub-patches as 16 and 4 respectively and used the linear combination of $R_{acc}$ (detection recall) and $R_{cost}$ which combines image acquisition cost and run-time performance reward.

\begin{spacing}{1}
\footnotesize
\begin{longtable}{|p{0.1\linewidth}|p{0.03\linewidth}|p{0.08\linewidth}|p{0.09\linewidth}|p{0.16\linewidth}|p{0.07\linewidth}|p{0.16\linewidth}|p{0.13\linewidth}|} 
\caption{Comparing various DRL-based object detection methods} 

\label{tab:obs}
\\
\hline
Approaches &
  Year &
  Training Technique &
  Actions &
  Remarks &
  Backbone &
  Performance &
  Datasets and source code \\ \hline \hline
  Active Object Localization \cite{caicedo2015active} &
  2015 &
  DQN &
  8 actions: up, down, left, right, bigger, smaller, fatter, taller &
  States: feature vector of observed region and action history. Reward: Change in IOU. &
  5 layer pretrained CNN &
  Higher mAP as compared to methods that did not use region proposals like MultiBox \cite{erhan2014scalable}, RegionLets \cite{zou2014generic}, DetNet \cite{szegedy2013deep}, and second best mAP as compared to R-CNN \cite{girshick2014rich} &
  Pascal VOC-2007 \cite{everingham2007pascal}, 2012 \cite{everingham2011pascal} Image Dataset.\\ \hline
Hierarchical Object Detection \cite{bellver2016hierarchical} &
  2016 &
  DQN &
  5 actions: 1 action per image quarter and 1 at the center &
  States: current 
  region and memory vector using Image-zooms and Pool45-crops. Reward: change in IOU. &
  VGG-16 \cite{simonyan2014very} &
  Objects detected with very few region proposals per image &
  Pascal VOC-2007 Image Dataset \cite{everingham2007pascal}. \newline \href{https://github.com/imatge-upc/detection-2016-nipsws}{\textcolor{blue}{\underline{Available Code}}} \\ \hline
Visual Object Detection \cite{mathe2016reinforcement} &
  2016 &
  Policy sampling and state transition algorithm &
  2 actions: fixate and done, where each is a tuple of three. &
  States: Observed region history, evidence region history and fixate history. Reward: sensitive to detection location &
  Deep NN \cite{krizhevsky2012imagenet} &
  Comparable mAP and lower run time as compared to other methods such as to exhaustive sliding window search(SW), exhaustive search over the CPMC and region proposal set(RP) \cite{gonzalez2015active} \cite{uijlings2013selective} &
  Pascal VOC 2012 Object detection challenge \cite{everingham2011pascal}. \\ \hline
  
  Tree-Structured Sequential Object Localization (Tree-RL) \cite{jie2016tree} &
  2016 &
  DQN &
  13 actions: 8 translation, 5 scaling. &
  States: Feature vector of current region, and whole image. Reward: change in IOU. &
  CNN trained on ImageNet \cite{deng2009imagenet} \cite{russakovsky2015imagenet} &
  Tree-RL with faster R-CNN outperformed RPN with fast R-CNN \cite{girshick2015fast} in terms of AP and comparable results to Faster R-CNN \cite{ren2015faster} &
  Pascal VOC 2007 \cite{everingham2007pascal} and 2012 \cite{everingham2011pascal}. \\ \hline
  Active Breast Lesion Detection \cite{maicas2017deep} &
  2017 &
  DQN &
  9 actions: 6 translation, 2 scaling, 1 trigger &
  States: feature vector of current region, Reward: improvement in localization. &
  ResNet \cite{he2016deep} &
 Comparable true positive and false positive proportions as compared to SL \cite{mcclymont2014fully} and Ms-C \cite{gubern2015automated}, but with lesser mean inference time. &
  DCE-MRI and T1-weighted anatomical dataset \cite{mcclymont2014fully} \\ \hline
  Multitask object localization \cite{wang2018multitask} &
  2018 &
  DQN &
  8 actions: left, right, up, down, bigger, smaller, fatter and taller &
  States: feature vector, historical actions. Reward: change in IOU. different network for transformation actions and terminal actions. &
  Pretrained VGG-16 \cite{simonyan2014very} with ImageNet \cite{deng2009imagenet} \cite{russakovsky2015imagenet} &
  Better mAP as compared to MultiBox \cite{erhan2014scalable}, Caicedo et al. \cite{caicedo2015active} and second best to R-CNN \cite{girshick2014rich}. &
  Pascal VOC-2007 Image Dataset \cite{everingham2007pascal}. \\ \hline
Bounding-Box Automated Refinement \cite{aylebar} &
  2020 &
  DQN &
  8 actions: up, down, left, right, bigger, smaller, fatter, taller &
  Offline and online implementation States: feature vector for offline (BAR-DRL), HOG for online (BAR-CB). Reward: change in IOU &
  ResNet50 \cite{he2016deep} &
  Better final IOU for boxes generated by methods such as RetinaNet \cite{lin2017focal}. &
  Pascal VOC-2007 \cite{everingham2007pascal}, 2012 \cite{everingham2011pascal} Image Dataset. \\ \hline
Efficient Object Detection in Large Images \cite{uzkent2020efficient} &
  2020 &
  DQN &
  binary action array: where 1 means that the agent would consider acquiring sub-patches for that particular patch &
  Course CPNet and fine FPNet level search. States: selected region. Reward: detection recall image acquisition cost. Policy: REINFORCE \cite{sutton2018reinforcement} &
  ResNet32 \cite{he2016deep} for policy network. and YOLOv3 \cite{redmon2018yolov3} with DarkNet-53 for Object detector &
  Higher mAP and lower run time as compared to other methods such as \cite{gao2018dynamic}. &
  Caltech  Pedestrian dataset (CPD) \cite{dollar2009pedestrian} \newline \href{https://github.com/uzkent/EfficientObjectDetection}{\textcolor{blue}{\underline{Available Code}}}\\ \hline

Organ Localization in CT \cite{navarro2020deep} &
  2020 &
  DQN &
  11 actions: 6 translation, 2 scaling, 3 deformation &
  States: region inside the Bounding box. Reward: change in IOU. &
  Architecture similar to \cite{alansary2019evaluating} &
  Lower distance error for organ localization and run time as compared to other methods such as 3D-RCNN \cite{xu2019efficient} and CNNs \cite{humpire2018efficient} & CT scans from the VISCERAL dataset \cite{jimenez2016cloud} \\ \hline

  Monocular 3D Object Detection \cite{Liu_ECCV2020_RL} & 2020 & DQN \cite{mnih2015human} & 15 actions, each modifies the 3D bounding box in a specific parameter & State: 3D bounding box parameters, 2D image of object cropped by 2D its detected bounding box. Reward: accuracy improvement after applying an action. & ResNet-101 \cite{he2016deep} & Higher average precision (AP) compared to \cite{3DBoundingBox_CVPR2017}, \cite{MonoGRNet_AAAI2019}, \cite{GS3D_CVPR2019} and \cite{M3DRPN_ICCV2019} & KITTI \cite{KITTI_dataset} \\ \hline

\end{longtable}
\end{spacing}

Localization of organs in CT scans is an important pre-processing requirement for taking the images of an organ, planning radiotherapy, etc. A DRL method for organ localization was proposed by \cite{navarro2020deep}, where the problem was formulated as an MDP. In the implementation, the agent was responsible for predicting a 3D bounding box around the organ. The authors used the last 4 states as input to the agent to stabilize the search and the action space consists of Eleven actions, 6 for the position of the bounding box, 2 for zoom in and zoom out the action, and last 3 for height, width, and depth. For Reward, they used the change the in Intersection over union (IOU) across an action.

Monocular 3D object detection is a problem where 3D bounding boxes of objects are required to be detected from a single 2D image. Even the sampling-based method is the SOTA approach, it has a huge flaw, in which most of the samples it generates do not overlap with the groundtruth. To leverage that method, \cite{Liu_ECCV2020_RL} introduced Reinforced Axial Refinement Network (RARN) for monocular 3D object detection by utilizing an RL model to iteratively refining the sampled bounding box to be more overlapped with the groundtruth bounding box. Given a state having the coordinates of the 3D bounding box and image patch of the image, the model predicts an action out of a set of 15 actions to refine one of the bounding box coordinates in a direction at every timestep, the model is trained by DQN method with the immediate reward is the improvement in detection accuracy between every pair of timesteps. The whole pipeline, namely RAR-Net, was evaluated on the real-world KITTI dataset \cite{KITTI_dataset} and achieved state-of-the-art performance.

\begin{figure}[!h]
	\centering \includegraphics[width=1.0\columnwidth]{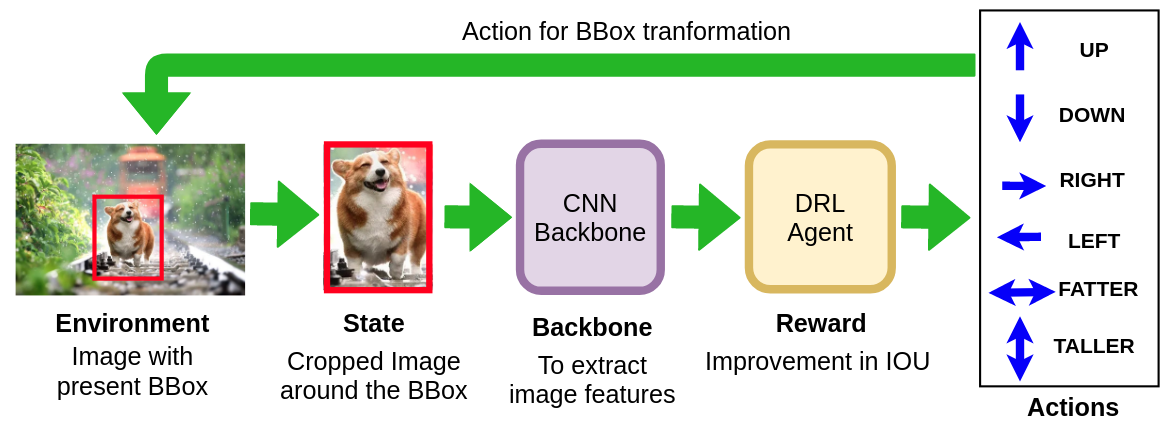}
		\caption{DRL implementation for object detection. The red box corresponds to the initial bounding box which for t=0 is predicted by some other algorithm or the transformed bounding box by previous iterations of DRL using the actions to maximize the improvement in IOU.}
	\label{fig:bbox}
\end{figure}

All these methods have been summarised and compared in Table \ref{tab:obs}, and a basic implementation of object detection using DRL has been shown in Fig. \ref{fig:bbox}. The figure illustrates a general implementation of object detection using DRL, where the state is an image segment cropped using a bounding box produced by some other algorithm or previous iteration of DRL, actions predicted by the DRL agent predict a series of bounding box transformation to fit the object better, hence forming a new state and Reward is the improvement in Intersection over union (IOU) with iterations as used by \cite{caicedo2015active},\cite{bellver2016hierarchical},\cite{aylebar},\cite{wang2018multitask},\cite{jie2016tree},\cite{navarro2020deep}.

\section{DRL in Object Tracking}
\label{sec:objecttracking}

Real-time object tracking has a large number of applications in the field of autonomous driving, robotics, security, and even in sports where the umpire needs accurate estimation of ball movement to make decisions. Object tracking can be divided into two main categories: Single object tracking (SOT) and Multiple object tracking (MOT).

Many attempts have been made for both SOT and MOT. SOT can be divided into two types, active and passive. In passive tracking it is assumed that the object that is being tracked is always in the camera frame, hence camera movement is not required. In active tracking, however, the decision to move the camera frame is required so that the object is always in the frame. Passive tracking has been performed by \cite{wu2013online}, \cite{hu2012single}, where \cite{hu2012single} performed tracking for both single and multiple objects. The authors of these papers proposed various solutions to overcome common problems such as a change in lighting and occlusion. Active tracking is a little bit harder as compared to a passive one because additional decisions are required for camera movement. Some efforts towards active tracking include \cite{denzler1994active} \cite{murray1994motion} \cite{kim2005detecting}. These solutions treat object detection and object tracking as two separate tasks and tend to fail when there is background noise. 

An end-to-end active object tracker using DRL was proposed by \cite{luo2017end}, where the authors used CNNs along with an LSTM \cite{hochreiter1997long} in their implementation. They used the actor-critic algorithm \cite{mnih2016asynchronous} to calculate the probability distribution of different actions and the value of state and used the object orientation and distance from the camera to calculate rewards. For experiments, the authors used VizDoom and Unreal Engine as the environment.

Another end-to-end method for SOT using sequential search strategy and DRL was proposed by \cite{zhang2017deep}. The method included using an RNN along with REINFORCE \cite{williams1992simple} algorithm to train the network. The authors used a function $f(W_{0})$ that takes in $S_{t}$ and frame as input, where $S_{t}$ is the object location for the first frame and is zero elsewhere. The output is fed to an LSTM module \cite{hochreiter1997long} with past hidden state $h_{t}$. The authors calculated the reward function by using insertion over union (IoU) and the difference between the average and max. 

A deformable face tracking method that could predict bounding box along with facial landmarks in real-time was proposed by \cite{guo2018dual}. The dual-agent DRL method (DADRL) mentioned in the paper consisted of two agents: a tracking and an alignment agent. The problem of object tracking was formulated as an MDP where state consisted of image regions extracted by the bounding box and a total of 8 actions (left, right, up, down, scale-up, scale down, stop and continue) were used, where first six consists of movement actions used by tracking agent and last two for alignment agent. The tracking agent is responsible for changing the current observable region and the alignment agent determines whether the iteration should be terminated. For the tracking agent, the reward corresponded to the misalignment descent and for the alignment agent the reward was $+\eta$ for misalignment less than the threshold and $-\eta$ otherwise. The DADRL implementation also consisted of communicated message channels beside the tracking agent and the alignment agent. The tracking agent consisted of a VGG-M \cite{simonyan2014very} backbone followed by a one-layer Q-Network and the alignment agent was designed as a combination of a stacked hourglass network with a confidence network. The two communicated message channels were encoded by a deconvolution layer and an LSTM unit \cite{hochreiter1997long} respectively.

Visual object tracking when dealing with deformations and abrupt changes can be a challenging task. A DRL method for object tracking with iterative shift was proposed by \cite{ren2018deep}. The approach (DRL-IS) consisted of three networks: The actor network, the prediction network, and the critic network, where all three networks shared the same CNN and a fully connected layer. Given the initial frame and bounding box, the cropped frame is fed to the CNNs to extract the features to be used as a state by the networks. The actions included continue, stop and update, stop and ignore, and restart. For continue, the bounding boxes are adjusted according to the output of the prediction network, for stop and update the iteration is stopped and the appearance feature of the target is updated according to the prediction network, for stop and ignore the updating of target appearance feature is ignored and restart means that the target is lost and the algorithm needs to start from the initial bounding box. The authors of the paper used reward as 1 for change in IoU greater than the threshold, 0 for change in IOU between + and - threshold, and -1 otherwise.

Considering the performance of actor-critic framework for various applications, \cite{chen2018real} proposed an actor-critic \cite{mnih2016asynchronous} framework for real-time object tracking. The authors of the paper used a pre-processing function to obtain an image patch using the bounding box that is fed into the network to find the bounding box location in subsequent frames. For actions the authors used $\triangle x$ for relative horizontal translation, $\triangle y$ for relative vertical translation, and $\triangle s$ for relative scale change, and for a reward they used 1 for IoU greater than a threshold and -1 otherwise. They proposed offline training and online tracking, where for offline training a pre-trained VGG-M \cite{simonyan2014very} was used as a backbone, and the actor-critic network was trained using the DDPG approach \cite{lillicrap2015continuous}. 

An improvement to \cite{chen2018real} for SOT was proposed by \cite{dunnhofer2019visual}, where a visual tracker was formulated using DRL and an expert demonstrator. The authors treated the problem as an MDP, where the state consists of two consecutive frames that have been cropped using the bounding box corresponding to the former frame and used a scaling factor to control the offset while cropping. The actions consisted of four elements: $\triangle x$ for relative horizontal translation, $\triangle y$ for relative vertical translation, $\triangle w$ for width scaling, and $\triangle h$ for height scaling, and the reward was calculated by considering whether the IoU is greater than a threshold or not. For the agent architecture the authors used a ResNet-18 \cite{he2016deep} as backbone followed by an LSTM unit \cite{wickelgren1973long}\cite{hochreiter1997long} to encode past information, and performed training based on the on-policy A3C framework \cite{mnih2016asynchronous}. 

In MOT the algorithm is responsible to track trajectories of multiple objects in the given video. Many attempts have been made with MOT including \cite{choi2015near}, \cite{chu2017online} and \cite{hong2016online}. However, MOT is a challenging task because of environmental constraints such as crowding or object overlapping. MOT can be divided into two main techniques: Offline \cite{choi2015near} and Online \cite{chu2017online} \cite{hong2016online}. In offline batch, tracking is done using a small batch to obtain tracklets and later all these are connected to obtain a complete trajectory. The online method includes using present and past frames to calculate the trajectory. Some common methods include Kalman filtering \cite{kim2014data}, Particle Filtering \cite{okuma2004boosted} or Markov decision \cite{xiang2015learning}. These techniques however are prone to errors due to environmental constraints.

To overcome the constraints of MOT by previous methods, \cite{xiang2015learning} proposed a method for MOT where the problem was approached as an MDP. The authors tracked each object in the frame through the Markov decision process, where each object has four states consisting: Active, Tracked, Lost, and Inactive. Object detection is the active state and when the object is in the lost state for a sufficient amount of time it is considered Inactive, which is the terminal state. The reward function in the implementation was learned through data by inverse RL problem \cite{ng2000algorithms}.

Previous approaches for MOT follow a tracking by detection technique that is prone to errors. An improvement was proposed by \cite{ren2018collaborative}, where detection and tracking of the objects were carried out simultaneously. The authors used a collaborative Q-Network to track trajectories of multiple objects, given the initial position of an object the algorithm tracked the trajectory of that object in all subsequent frames. For actions the authors used $\triangle x$ for relative horizontal translation, $\triangle y$ for relative vertical translation, $\triangle w$ for width scaling, and $\triangle h$ for height scaling, and the reward consisted of values 1,0,-1 based on the IoU. 

Another method for MOT was proposed by \cite{jiang2018multiobject}, where the authors used LSTM \cite{hochreiter1997long} and DRL to approach the problem of multi-object tracking. The method described in the paper used three basic components: a YOLO V2 \cite{milan2017online} object detector, many single object trackers, and a data association module. Firstly the YOLO V2 object detector is used to find objects in a frame, then each detected object goes through the agent which consists of CNN followed by an LSTM to encode past information for the object. The state consisted of the image patch and history of past 10 actions, where six actions (right, left, up, down, scale-up, scale down) were used for bounding box movement across the frame with a stop action for the terminal state. To provide reinforcement to the agent the reward was 1 if the IOU is greater than a threshold and 0 otherwise. In their experiments, the authors used VGG-16 \cite{simonyan2014very} for CNN backbone and performed experiments on MOT benchmark \cite{leal2015motchallenge} for people tracking.
\begin{spacing}{1.2}
\footnotesize
\begin{longtable}{|p{0.11\linewidth}|p{0.04\linewidth}|p{0.08\linewidth}|p{0.09\linewidth}|p{0.15\linewidth}|p{0.09\linewidth}|p{0.15\linewidth}|p{0.13\linewidth}|} 
\caption{Comparing various DRL-based object tracking methods. The First group for Single object tracking (SOT) and the second group for multi-object tracking (MOT)} 

\label{tab:track}
\\
\hline
Approaches &
  Year &
  Training Technique &
  Actions &
  Remarks &
  Backbone &
  Performance &
  Datasets and Source code \\ \hline \hline
End to end active object tracking \cite{luo2017end} &
  2017 &
  Actor-Critic (a3c) \cite{mnih2016asynchronous} &
  6 actions: turn left, turn right, turn left and move forward, turn right and move forward, move forward, no-op &
  Environment: virtual environment. Reward: calculated using object orientation and position. Tracking Using LSTM \cite{hochreiter1997long}\ &
  ConvNet-LSTM &
   Higher accumulated reward and episode length as compared to methods like MIL \cite{babenko2009visual}, Meanshift \cite{comaniciu2000real}, KCF \cite{henriques2014high}. &
  ViZDoom \cite{kempka2016vizdoom}, Unreal Engine \\ \hline
DRL for object tracking \cite{zhang2017deep} &
  2017 &
  DRLT &
  None &
  State: feature vector, Reward: change in IOU use of LSTM \cite{hochreiter1997long} and REINFORCE \cite{williams1992simple} &
  YOLO network \cite{redmon2016you} &
  Higher area under curve (success rate Vs overlap threshold), precision and speed (fps) as compared to STUCK \cite{hare2015struck} and DLT \cite{wang2013learning}. &
  Object tracking benchmark \cite{wu2013online}. \newline \href{https://github.com/fgabel/Deep-Reinforcement-Learning-for-Visual-Object-Tracking-in-Videos}{\textcolor{blue}{\underline{Available Code}}} \\ \hline
 Dual-agent deformable face tracker \cite{guo2018dual} &
  2018 &
  DQN &
  8 actions: left, right, up, down, scale up, scale down, stop and continue. &
  States: image region using Bounding box. Reward: distance error. Facial landmark detection and tracking using LSTM \cite{hochreiter1997long} &
  VGG-M \cite{simonyan2014very} &
   Lower normalized point to point error for landmarks and higher success rate for facial tracking as compared to ICCR \cite{krizhevsky2012imagenet}, MDM \cite{shen2015first}, Xiao et al \cite{black1995tracking}, etc. &
  Large-scale face tracking dataset, the 300-VW test set \cite{shen2015first} \\ \hline 
  Tracking with iterative shift \cite{ren2018deep} &
  2018 &
  Actor-critic \cite{mnih2016asynchronous} &
  4 actions: continue, stop and update, stop and ignore and restart &
  States: image region using bounding box. Reward: change in IOU. Three networks: actor, critic and prediction network &
   3 Layer CNN and FC layer &
   Higher area under curve for success rate Vs overlap threshold and precision Vs location error threshold as compared to CREST \cite{song2017crest}, ADNet \cite{yun2017action}, MDNet \cite{nam2016learning}, HCFT \cite{ma2015hierarchical}, SINT \cite{tao2016siamese}, DeepSRDCF \cite{danelljan2015learning}, and HDT \cite{qi2016hedged} &
  OTB-2015  \cite{wu2015object}, Temple-Color \cite{liang2015encoding}, and VOT-2016 Dataset \cite{kristan2015visual} \\ \hline
Tracking with actor-critic \cite{chen2018real} &
  2018 &
  Actor-critic \cite{mnih2016asynchronous} &
  3 actions: $\triangle x$, $\triangle y$ and $\triangle s$ &
  States: image region using bounding box. Reward: IOU greater then threshold. Offline training, online tracking &
  VGG-M \cite{simonyan2014very} &
  Higher average precision score then PTAV \cite{fan2017parallel},  CFNet \cite{valmadre2017end}, ACFN \cite{choi2017attentional}, SiameFC \cite{bertinetto2016fully}, ECO-HC \cite{danelljan2015learning}, etc. &
  OTB-2013 \cite{wu2013online}, OTB-2015 \cite{wu2015object} and VOT-2016 dataset \cite{kristan2015visual} \newline \href{https://github.com/bychen515/ACT}{\textcolor{blue}{\underline{Available Code}}} \\ \hline
Visual tracking and expert  demonstrator \cite{dunnhofer2019visual} &
  2019 &
  Actor-critic (a3c) \cite{mnih2016asynchronous} &
   4 actions:  $\triangle x$, $\triangle y$,$\triangle w$ and $\triangle h$ &
   States: image region using bounding box. Reward: change in IOU. SOT using LSTM \cite{wickelgren1973long}\cite{hochreiter1997long} &
  ResNet-18 \cite{he2016deep} &
   Comparable success and precision scores as compared to LADCF \cite{xu2019learning}, SiamRPN \cite{li2018high} and ECO \cite{danelljan2017eco} &
   GOT-10k \cite{huang2019got}, LaSOT \cite{fan2019lasot}, UAV123 \cite{mueller2016benchmark}, OTB-100 \cite{wu2013online}, VOT-2018 \cite{kristan2018sixth} and VOT-2019. \\ \hline \hline
  
   Object tracking by decision making \cite{xiang2015learning} &
  2015 & 
  TLD Tracker \cite{kalal2011tracking} &
   7 actions: corresponding to moving the object between states such as Active, tracked, lost and Inactive &
   States: 4 states: Active, tracked, lost and Inactive. Reward: inverse RL problem \cite{ng2000algorithms} &
  None &
   Comparable multiple object tracking accuracy (MOTA) and multiple object tracking precision (MOTP) \cite{bernardin2008evaluating} as compared to DPNMS \cite{pirsiavash2011globally}, TCODAL \cite{bae2014robust}, SegTrack \cite{milan2015joint}, MotiCon \cite{leal2014learning}, etc &
   M0T15 dataset \cite{leal2015motchallenge} \newline \href{https://github.com/yuxng/MDP_Tracking}{\textcolor{blue}{\underline{Available Code}}}\\ \hline
 Collaborative multi object tracker \cite{ren2018collaborative} &
  2018 &
  DQN &
  4 actions: $\triangle x$,  $\triangle y$, $\triangle w$ and $\triangle h$ &
   States: image region using bounding box. Reward: IOU greater then threshold. 2 networks: prediction and decision network &
   3 Layer CNN and FC Layer &
  Comparable multiple object tracking accuracy (MOTA) and multiple object tracking precision (MOTP) \cite{bernardin2008evaluating} as compared to SCEA \cite{hong2016online}, MDP \cite{xiang2015learning}, CDADDALpb \cite{bae2017confidence}, AMIR15 \cite{sadeghian2017tracking} &
 MOT15  \cite{leal2015motchallenge} and MOT16 \cite{milan2016mot16} datasets \\ \hline

 Multi object tracking in video \cite{jiang2018multiobject} &
  2018 &
  DQN &
  6 actions: right, left, up, down, scale up, scale down &
  States: image region using bounding box. Reward: IOU greater then threshold. Detection using YOLO-V2 \cite{milan2017online} for detector and LSTM \cite{hochreiter1997long} . &
  VGG-16 \cite{simonyan2014very} &
  Comparable if not better multiple object tracking accuracy (MOTA) and multiple object tracking precision (MOTP) \cite{bernardin2008evaluating} as compared to RNN-LSTM \cite{leal2015motchallenge}, LP-SSVM \cite{xiang2015learning}, MDPSubCNN \cite{leal2016learning}, and SiameseCNN \cite{hamid2015joint} &
  MOT15 Dataset \cite{leal2015motchallenge} \\ \hline
  Multi agent multi object tracker \cite{jiang2019multi} &
  2019 &
  DQN &
  9 actions: move right, move left, move up, move down, scale up, scale down, fatter, taller and stop &
  States: image region using bounding box. Reward: IOU greater then threshold. YOLO-V3 \cite{redmon2018yolov3} for detection and LSTM \cite{hochreiter1997long}. &
  VGG-16 \cite{simonyan2014very} &
  Higher running time, and comparable if not better multiple object tracking accuracy (MOTA) and multiple object tracking precision (MOTP) \cite{bernardin2008evaluating} as compared to RNN-LSTM \cite{leal2015motchallenge}, LP-SSVM \cite{xiang2015learning}, MDPSubCNN \cite{leal2016learning}, and SiameseCNN \cite{hamid2015joint} &
  MOT15 challenge benchmark  \cite{leal2015motchallenge}. \\ \hline

\end{longtable}
\end{spacing}

To address the problems in existing tracking methods such as varying numbers of targets, non-real-time tracking, etc, \cite{jiang2019multi} proposed a multi-object tracking algorithm based on a multi-agent DRL tracker (MADRL). In their object tracking pipeline the authors used YOLO-V3 \cite{redmon2018yolov3} as object detector, where multiple detections produced by YOLO-V3 were filtered using the IOU and the selected results were used as multiple agents in multiple agent detector. The input agents were fed into a pre-trained VGG-16 \cite{simonyan2014very} followed by an LSTM unit \cite{hochreiter1997long} that could share information across agents and return the actions encoded in a 9-dimensional vector( move right, move left, move up, move down, scale-up, scale down, aspect ratio change fatter, aspect ratio change taller and stop), also a reward function similar to \cite{jiang2018multiobject} was used. 

\begin{figure}[!h]
	\centering \includegraphics[width=1.0\columnwidth]{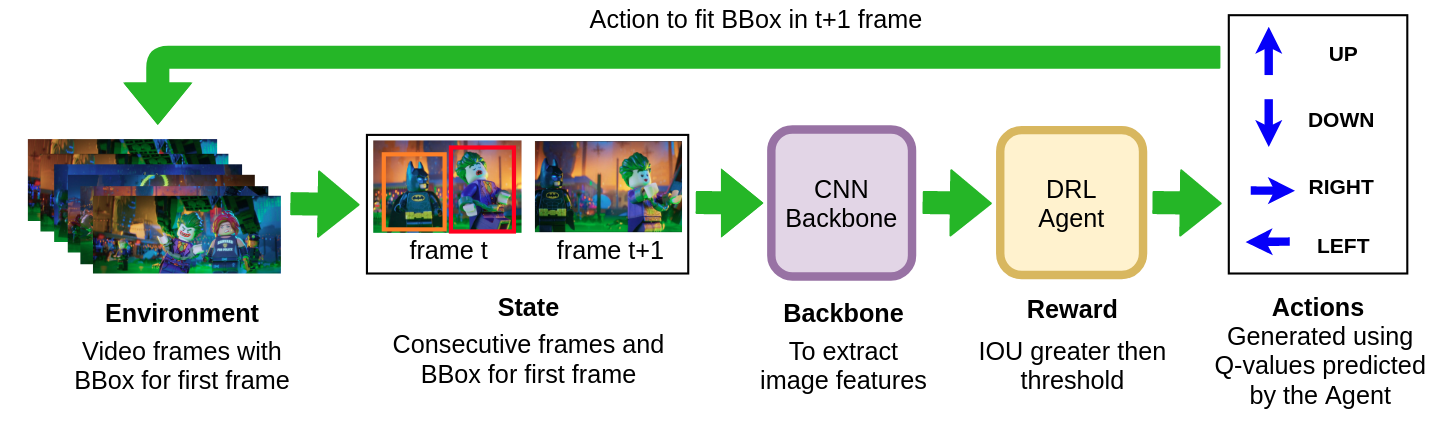}
		\caption{DRL implementation for object tracking. Here the state consists of two consecutive frames with bounding box locations for the first frame that is predicted by some object detection algorithm or by the previous iteration of DRL, the actions move the bounding box present in the first frame to fit the object in the second frame to maximize the reward which is the whether the IOU is greater than a given threshold or not.}
	\label{fig:track}
\end{figure}

Various works in the field of object tracking have been summarized in Table \ref{tab:track}, and a basic implementation of object tracking using DRL has been shown in Fig. \ref{fig:track}. The figure illustrates a general implementation of object tracking in videos using DRL, where the state consists of two consecutive frames $(F_{t},F_{t+1})$ with a bounding box for the first frame produced by another algorithm for the first iteration or by the previous iterations of DRL agent. The actions corresponds to the moving the bounding on the image to fit the object in frame $F_{t+1}$, hence forming a new state with frame $F_{t+1}$ and frame $F_{t+2}$ along with the bounding box for frame $F_{t+1}$ predicted by previous iteration and reward corresponds to whether IOU is greater then a given threshold as used by \cite{guo2018dual},\cite{ren2018deep},\cite{chen2018real}, \cite{dunnhofer2019visual},\cite{ren2018collaborative},\cite{jiang2018multiobject},\cite{jiang2019multi}.    

\section{DRL in Image Registration}
\label{sec:imageregistration}
Image registration is a very useful step that is performed on 3D medical images for the alignment of two or more images. The goal of 3D medical image registration is to find a correlation between two images from either different patients or the same patients at different times, where the images can be Computed Tomography (CT), Magnetic Resonance Imaging (MRI), or Positron Emission Tomography (PET). In the process, the images are brought to the same coordinate system and aligned with each other. The reason for image registration being a challenging task is the fact that the two images used may have a different coordinate system, scale, or resolution.

Many attempts have been made toward automated image registration. A multi-resolution strategy with local optimizers to perform 2D or 3D image registration was performed by \cite{thevenaz2000optimization}. However, multi-resolution tends to fail with different field of views. Heuristic semi-global optimization schemes were proposed to solve this problem and used by  \cite{matsopoulos1999automatic} through simulated annealing and through genetic algorithm \cite{rouet2000genetic}, However, their cost of computation was very high. A CNN-based approach to this problem was suggested by \cite{miao2016cnn}, and \cite{dosovitskiy2015flownet} proposed an optical flow method between 2D RGB images. A descriptor learned through a CNN was proposed by \cite{wohlhart2015learning}, where the authors encoded the posture and identity of a 3D object using the 2D image. Although all of these formulations produce satisfactory results yet, the methods could not be applied directly to 3D medical images.

To overcome the problems faced by previous methods, \cite{lotfi2013improving} proposed a method for improving probabilistic image registration via RL and uncertainty evaluation. The method involved predicting a regression function that predicts registration error from a set of features by using regression random forests (RRF) \cite{breiman1996bagging} method for training. The authors performed experiments on 3D MRI images and obtained an accuracy improvement of up to 25\%.

Previous image registration methods are often customized to a specific problem and are sensitive to image quality and artifacts. To overcome these problems, \cite{liao2017artificial} proposed a robust method using DRL. The authors considered the problem as an MDP where the goal is to find a set of transformations to be performed on the floating image to register it on the reference image. They used the gamma value for future reward decay and used the change in L2 Norm between the predicted transformation and ground truth transformation to calculate the reward. The authors also used a hierarchical approach to solve the problem with varying FOVs and resolutions.

\begin{spacing}{1.2}
\footnotesize
\begin{longtable}{|p{0.11\linewidth}|p{0.04\linewidth}|p{0.08\linewidth}|p{0.09\linewidth}|p{0.15\linewidth}|p{0.09\linewidth}|p{0.15\linewidth}|p{0.13\linewidth}|}
\caption{Comparing various DRL-based image registration methods.} 

\label{tab:reg}
\\
\hline
 Approaches &
  Year &
   Training Technique &
  Actions &
  Remarks &
  Backbone &
  Performance &
  Datasets \\ \hline \hline
   Image registration using uncertainity evaluation \cite{lotfi2013improving} &
  2013 &
  DQN &
  Not specified &
  Probabilistic model using regression random forests (RRF) \cite{breiman1996bagging} &
  Not specified &
  Higher final Dice score (DSC) as compared to other methods like random seed selection and grid-based seed selection &
  3D MRI images from LONI Probabilistic Brain Atlas (LPBA40)  \href{http://www.loni.ucla.edu/}{Dataset} \\ \hline
 Robust Image registration \cite{liao2017artificial} &
  2017 &
  DQN &
  12 actions: corresponding to different transformations &
  States: current transformation. Reward: distance error. &
  5 Conv3D layers and 3 FC layers &
  Better success rate then ITK \cite{ibanez2005itk}, Quasi-global \cite{miao2013system} and Semantic registration\cite{neumann2014probabilistic} &
  Abdominal spine CT and CBCT dataset, Cardiac CT and CBCT \\ \hline

 Multimodal image registration \cite{ma2017multimodal} &
  2017 &
  Duel-DQN Double-DQN &
  Actions update the transformations on floating image &
 States: cropped 3D image. Duel-DQN for value estimation and Double DQN for updating weights. &
   Batch normalization followed by 5 Conv3D and 3 Maxpool layers &
   Lower Euclidean distance error as compared to methods like Hausdorff, ICP, DQN \cite{mnih2015human}, Dueling \cite{wang2015dueling}, etc. &
   Thorax and Abdomen (ABD) dataset \\ \hline
 Robust non-rigid agent-based registration \cite{krebs2017robust} &
  2017 &
  DQN &
 2n actions for n dimensional $\theta$ vector &
   States: fixed and moving image. Reward: change in transformation error. With Statistical deformation model and fuzzy action  control. &
   Multi layer CNN, pooling and FC layers. &
   Higher Mean Dice score and lower Hausdorff distance as compared to methods like LCC-Demons \cite{lorenzi2013lcc} and Elastix \cite{klein2009elastix}. &
   MICCAI challenge PROMISE12 \cite{litjens2014evaluation} \\ \hline
  
   Robust Multimodal registration \cite{sun2018robust} &
  2018 &
  Actor-Critic (a3c) \cite{mnih2016asynchronous} &
  8 actions: for different transformations &
   States: fixed and moving image. Reward: Distance error. Monte-carlo method with LSTM \cite{hochreiter1997long}. &
   Multi layer CNN and FC layer &
   Comparable if not lower target registration error \cite{fitzpatrick2001distribution} as compared to methods like SIFT \cite{lowe2004distinctive}, Elastix \cite{klein2009elastix}, Pure SL, RL-matrix, RL-LME, etc. &
  CT and MR images \\ \hline


\end{longtable}
\end{spacing}

A multi-modal method for image registration was proposed by \cite{ma2017multimodal}, where the authors used DRL for alignment of depth data with medical images. In the specified work Duel DQN was used as the agent for estimating the state value and the advantage function, and the cropped 3D image tensor of both data modalities was considered as the state. The algorithm's goal was to estimate a transformation function that could align moving images to a fixed image by maximizing a similarity function between the fixed and moving image. A large number of convolution and pooling layer were used to extract high-level contextual information, batch normalization and concatenation of feature vector from last convolution layer with action history vector was used to solve the problem of oscillation and closed loops, and Double DQN architecture for updating the network weights was used. 

Previous methods for image registration fail to cope with large deformations and variability in appearance. To overcome these issues \cite{krebs2017robust} proposed a robust non-rigid agent-based method for image registration. The method involves finding a spatial transformation $T_{\theta}$ that can map the fixed image with the floating image using actions at each time step, that is responsible for optimizing $\theta$. If the $\theta$ is a d dimensional vector then there will be 2d possible actions. In this work, a DQN was used as an agent for value estimation, along with a reward that corresponded to the change in $\theta$ distance between ground truth and predicted transformations across an action. 

An improvement to the previous methods was proposed by \cite{sun2018robust}, where the authors used a recurrent network with RL to solve the problem. Similar to \cite{liao2017artificial}, they considered the two images as a reference/fixed and floating/moving, and the algorithm was responsible for predicting transformation on the moving image to register it on a fixed image. In the specified work an LSTM \cite{hochreiter1997long} was used to encode past hidden states, Actor-critic \cite{mnih2016asynchronous} for policy estimation, and a reward function corresponding to distance between ground truth and transformed predicted landmarks were used. 

\begin{figure}[!h]
	\centering \includegraphics[width=1.0\columnwidth]{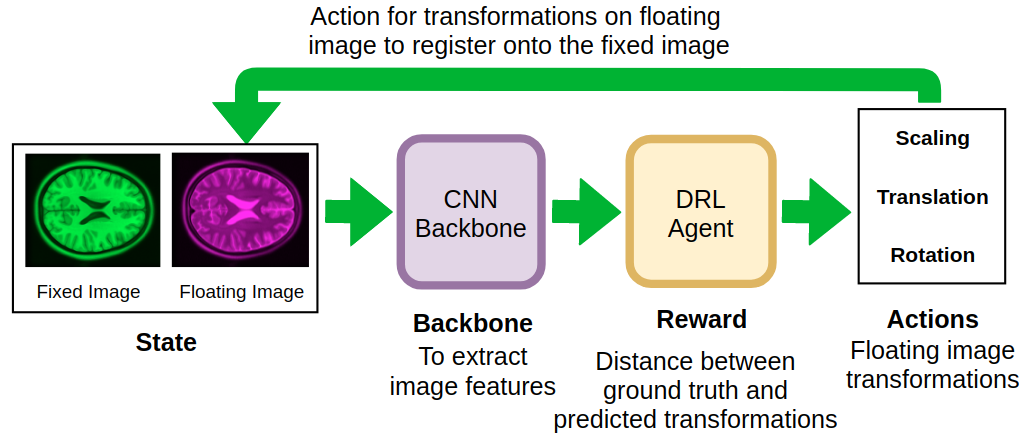}
		\caption{DRL implementation for image registration. The state consists of fixed and floating image and the actions in form of transformations are performed on the floating image so as to maximize reward by minimizing distance between the ground truth and predicted transformations. }
	\label{fig:reg}
\end{figure}

Various methods in the field of Image registration have been summarized and compared in Table \ref{tab:reg}, and a basic implementation of image registration using DRL has been shown in Fig. \ref{fig:reg}. The figure illustrates a general implementation of image registration using DRL where the state consists of a fixed and floating image. The DRL agent predicts actions in form of a set of transformations on a floating image to register it onto the fixed image hence forming a new state and accepts reward in form of improvement in distance error between ground truth and predicted transformations with iterations as described by \cite{sun2018robust},\cite{krebs2017robust},\cite{liao2017artificial}.

\section{DRL in Image Segmentation}
\label{sec:imagesegmentation}

Image segmentation is one of the most extensively performed tasks in computer vision, where the algorithm is responsible for labeling each pixel position as foreground or background corresponding to the object being segmented in the image. Image segmentation has a wide variety of applications in medical, robotics, weather, etc. One of the earlier attempts with image segmentation includes \cite{haralick1985image}. With the improvement in detection techniques and introduction of CNN, new methods are introduced every year for image segmentation. Mask R-CNN \cite{he2017mask} extended the work by Faster R-CNN \cite{ren2015faster} by adding a segmentation layer after the Bounding box has been predicted. Some earlier works include \cite{girshick2014rich}, \cite{hariharan2014simultaneous}, \cite{hariharan2015hypercolumns} etc. Most of these works give promising results in image segmentation. However, due to the supervised nature of CNN and R-CNN, these algorithms need a large amount of data. In fields like medical, the data is sometimes not readily available hence we needed a way to train algorithms to perform a given task when there are data constraints. Luckily RL tends to shine when the data is not available in a large quantity. 

One of the first methods for Image segmentation through RL was proposed by \cite{sahba2006reinforcement}, where the authors proposed an RL framework for medical image segmentation. In their work, they used a Q-Matrix, where the actions were responsible for adjusting the threshold values to predict the mask and the reward was the normalized change in quality measure between action steps. \cite{sahba2007application} also used a similar technique of Tabular method. 

To overcome the constraints of the previous method for segmentation, \cite{reza2016reinforcement} proposed a method for indoor semantic segmentation through RL. In their paper, the authors proposed a sequential strategy using RL to combine binary object masks of different objects into a single multi-object segmentation mask. They formulated the binary mask in a Conditional Random Field Framework (CRF), and used a logistic regression version of AdaBoost \cite{hoiem2007recovering} for classification. The authors considered the problem of adding multiple binary segmentation into one as an MDP, where the state consisted of a list of probability distributions of different objects in an image, and the actions correspond to the selection of object/background segmentation for a particular object in the sequential semantic segmentation. In the RL framework, the reward was considered in terms of pixel-wise frequency weighted Jaccard Index computed over the set of actions taken at any stage of an episode.

Interactive segmentation is the task of producing an interactive mask for objects in an image. Most of the previous works in this field greatly depend on the distribution of inputs which is user-dependent and hence produce inadequate results. An improvement was proposed by \cite{song2018seednet}, where the authors proposed SeedNet, an automatic seed generation method for robust interactive segmentation through RL. With the image and initial seed points, the algorithm is capable of generating additional seed points and image segmentation results. The implementation included Random Walk (RW) \cite{grady2006random} as the segmentation algorithm and DQN for value estimation by considering the problem as an MDP. They used the current binary segmentation mask and image features as the state, the actions corresponded to selecting seed points in a sparse matrix of size $20 \times 20$(800 different actions were possible), and the reward consisted of the change in IOU across an action. In addition, the authors used an exponential IOU model to capture changes in IOU values more accurately. 

Most of the previous work for image segmentation fail to produce satisfactory results when it comes to 3D medical data. An attempt on 3D medical image segmentation was done by  \cite{liao2020iteratively}, where the authors proposed an iteratively-refined interactive multi-agent method for 3D medical image segmentation. They proposed a method to refine an initial course segmentation produced by some segmentation methods using RL, where the state consisted of the image, previous segmentation probability, and user hint map. The actions corresponded to adjusting the segmentation probability for refinement of segmentation, and a relative cross-entropy gain-based reward to update the model in a constrained direction was used. In simple words, it is the relative improvement of previous segmentation to the current one. The authors utilized an asynchronous advantage actor-critic algorithm for determining the policy and value of the state.

\begin{spacing}{1.2}
\footnotesize
\begin{longtable}{|p{0.09\linewidth}|p{0.04\linewidth}|p{0.11\linewidth}|p{0.09\linewidth}|p{0.15\linewidth}|p{0.08\linewidth}|p{0.15\linewidth}|p{0.13\linewidth}|}
\caption{Comparing various DRL-based image segmentation methods} 

\label{tab:seg}
\\
\hline
Approaches &
  Year &
  Training Technique &
  Actions &
  Remarks &
  Backbone &
  Performance &
  Datasets \\ \hline
 Semantic Segmentation for indoor scenes\cite{reza2016reinforcement} &
  2016 &
  DQN &
  2 actions per object: object, background &
   States: current probability distribution. Reward: pixel-wise frequency weighted Jaccard index. Conditional Random Field Framework (CRF) and logistic regression version of AdaBoost \cite{hoiem2007recovering} for classification. &
  Not Specified &
  Pixel-wise percentage jaccard index comparable to Gupta-L \cite{gupta2014learning} and Gupta-P \cite{gupta2013perceptual}. &
  NYUD V2 dataset \cite{silberman2012indoor} \\ \hline
SeedNet \cite{song2018seednet} &
  2018 &
   DQN, Double-DQN, Duel-DQN &
  800 actions: 2 per pixel &
  States: image features and segmentation mask. Reward: change in IOU.  Random Walk (RW) \cite{grady2006random} for segmentation algorithm. &
  Multi layer CNN &
  Better IOU then methods like FCN \cite{long2015fully} and iFCN \cite{xu2016deep}. &
  MSRA10K saliency dataset \cite{cheng2014global} \\ \hline
 Iteratively refined multi agent  segmentation \cite{liao2020iteratively} &
  2020 &
  Actor-critic (a3c) \cite{mnih2016asynchronous} &
  1 action per voxel for adjusting segmentation probability &
   States: 3D image segmentation probability and hint map. Reward: cross entropy gain based framework. &
  R-net \cite{wang2018deepigeos} &
   Better performance then methods like MinCut \cite{krahenbuhl2011efficient}, DeepIGeoS (R-Net) \cite{wang2018deepigeos} and InterCNN \cite{bredell2018iterative}. &
   BraTS 2015\cite{menze2014multimodal}, MM-WHS \cite{zhuang2016multi} and  NCI-ISBI 2013 Challenge \cite{bloch2015nci} \\ \hline
  Multi-step medical image segmentation \cite{tian2020multi} &
  2020 &
  Actor-critic (a3c) \cite{mnih2016asynchronous} &
  Actions control the position and shape of brush stroke to modify segmentation &
  States: image, segmentation mask and time step. Reward: change in distance error. Policy: DPG \cite{silver2014deterministic}. &
  ResNet18 \cite{he2016deep} &
  Higher Mean Dice score and lower Hausdorff distance then methods like Grab-Cut \cite{rother2004grabcut}, PSPNet \cite{zhao2017pyramid}, FCN \cite{long2015fully}, U-Net \cite{ronneberger2015u}, etc. &
  Prostate MR image dataset (PROMISE12, ISBI2013) and retinal fundus  image dataset (REFUGE challenge dataset \cite{orlando2020refuge}) \\ \hline
  Anomaly Detection in Images \cite{Chu_RL_ECCV2020} & 2020 & REINFORCE \cite{williams1992simple} & 9 actions, 8 for directions to shift center of the extracted patch to, the last action is to switch to a random new image & Environment: input image to the model. State: observed patch from the image centered by predicted center of interest.
  & None & Superior performance in \cite{MVTecAD_CVPR2019} and \cite{CrackForest_TITS2016} on all metrics e.g. precision, recall and F1 when compared with U-Net \cite{ronneberger2015u} and baseline unsupervised method in \cite{MVTecAD_CVPR2019} but only wins on recall in \cite{NanoTWICE_TII2017} &
  MVTec AD \cite{MVTecAD_CVPR2019}, NanoTWICE \cite{NanoTWICE_TII2017}, CrackForest \cite{CrackForest_TITS2016} \\ \hline

\end{longtable}
\end{spacing}

Further improvement in the results of medical image segmentation was proposed by \cite{tian2020multi}. The authors proposed a method for multi-step medical image segmentation using RL, where they used a deep deterministic policy gradient method (DDPG) based on actor-critic framework \cite{mnih2016asynchronous} and similar to Deterministic policy gradient (DPG) \cite{silver2014deterministic}. The authors used ResNet18 \cite{he2016deep} as backbone for actor and critic network along with batch normalisation \cite{ioffe2015batch} and weight normalization with Translated ReLU \cite{xiang2017effects}.  In their MDP formulation, the state consisted of the image along with the current segmentation mask and step-index, and the reward corresponded to the change in mean squared error between the predicted segmentation and ground truth across an action. According to the paper the action was defined to control the position and shape of brush stroke used to modify the segmentation.


An example in image segmentation outside the medical field is \cite{Chu_RL_ECCV2020} proposing to tackle the problem of anomalies detection and segmentation in images (i.e. damaged pins of an IC chip, small tears in woven fabric). \cite{Chu_RL_ECCV2020} utilizes an additional module to attend only on a specific patch of the image centered by a predicted center instead of the whole image, this module helps a lot in reducing the imbalance between normal regions and abnormal locations. Given an image, this module, namely Neural Batch Sampling (NBS), starts from a random initiated center and recurrently moves that center by eight directions to the abnormal location in the image if it exists, and it has an additional action to stop moving the center when it has already converged to the anomaly location or there is not any anomaly can be observed. The NBS module is trained by REINFORCE algorithm \cite{williams1992simple} and the whole model is evaluated on multiple datasets e.g. MVTec AD \cite{MVTecAD_CVPR2019}, NanoTWICE \cite{NanoTWICE_TII2017}, CrackForest \cite{CrackForest_TITS2016}.

\begin{figure}[!h]
	\centering \includegraphics[width=1.0\columnwidth]{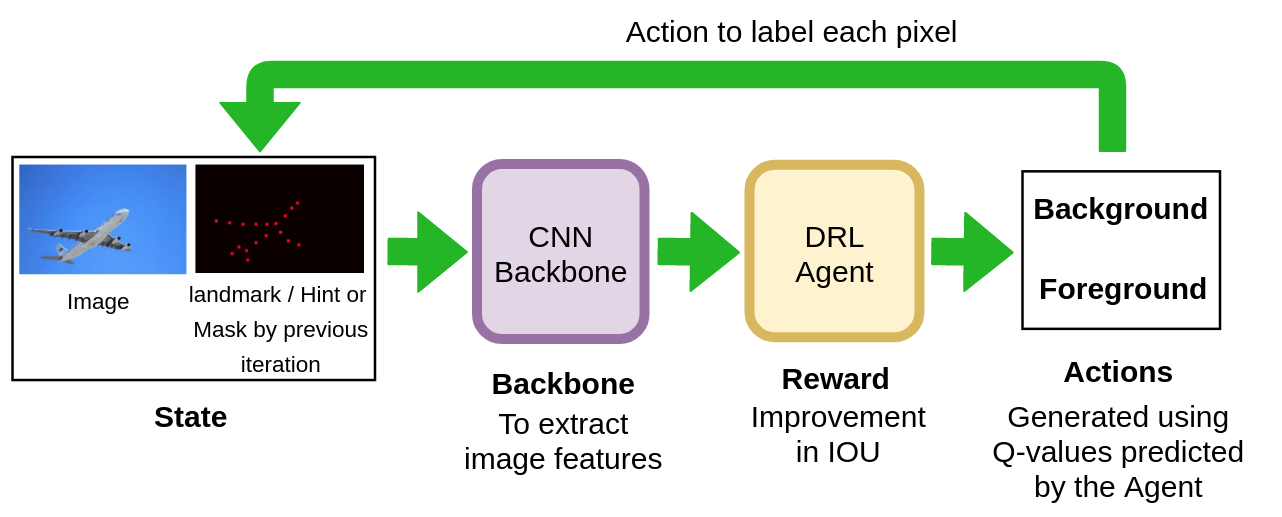}
		\caption{DRL implementation for Image segmentation. The state consists of the image to be segmented along with a user hint for t=0 or the segmentation mask by the previous iterations. The DRL agent performs actions by labeling each pixel as foreground and background to maximize the improvement in IOU over the iterations.}
	\label{fig:seg}
\end{figure}

Various works in the fields of Image segmentation have been summarised and compared in Table \ref{tab:seg},  and a basic implementation of image segmentation using DRL has been shown in Fig. \ref{fig:seg}. The figure shows a general implementation of image segmentation using DRL. The states consist of the image along with user hint (landmarks or segmentation mask by the other algorithm) for the first iteration or segmentation mask by the previous iteration. The actions are responsible for labeling each pixel as foreground and background and reward corresponds to an improvement in IOU with iterations as used by \cite{song2018seednet},\cite{liao2020iteratively}.

\section{DRL in Video Analysis}
\label{sec:videoapplications}

Object segmentation in videos is a very useful yet challenging task in computer vision field. Video object segmentation task focuses on labelling each pixel for each frame as foreground or background. Previous works in the field of video object segmentation can be divided into three main methods. unsupervised  \cite{papazoglou2013fast}\cite{xiao2016track}, weakly supervised \cite{cheng2017segflow}\cite{jain2017fusionseg} \cite{zhang2017spftn} and semi-supervised \cite{caelles2017one} \cite{jampani2017video}\cite{perazzi2017learning}. 

A DRL-based framework for video object segmentation was proposed by \cite{sahba2016deep}, where the authors divided the image into a group of sub-images and then used the algorithm on each of the sub-image. They proposed a group of actions that can perform to change the local values inside each sub-image and the agent received reward based on the change in the quality of segmented object inside each sub-image across an action. In the proposed method deep belief network (DBN) \cite{chen2015spectral} was used for approximating the Q-values.

Surgical gesture recognition is a very important yet challenging task in the computer vision field. It is useful in assessing surgical skills and for efficient training of surgeons. A DRL method for surgical gesture classification and segmentation was proposed by \cite{liu2018deep}. The proposed method could work on features extracted by video frames or kinematic data frames collected by some means along with the ground truth labels. The problem of classification and segmentation was considered as an MDP, where the state was a concatenation of TCN \cite{lea2017temporal}\cite{leal2016learning} features of the current frame, 2 future frames a specified number of frames later, transition probability of each gesture computed from a statistical language model \cite{richard2016temporal} and a one-hot encoded vector for gesture classes. The actions could be divided into two sub-actions, One to decide optimal step size and one for choosing gesture class, and the reward was adopted in a way that encouraging the agent to adopt a larger step and also penalizes the agent for errors caused by the action. The authors used  Trust Region Policy Optimization (TRPO) \cite{schulman2015trust} for training the policy and a spacial CNN \cite{lea2016segmental} to extract features.

Earlier approaches for video object segmentation required a large number of actions to complete the task. An Improvement was proposed by \cite{han2018reinforcement}, where authors used an RL method for object segmentation in videos. They proposed a reinforcement cutting-agent learning framework, where the cutting-agent consists of a cutting-policy network (CPN) and a cutting-execution network (CEN). The CPN learns to predict the object-context box pair, while CEN learns to predict the mask based on the inferred object-context box pair. The authors used MDP to solve the problem in a semi-supervised fashion. For the state of CPN  the authors used the input frame information, the action history, and the segmentation mask provided in the first frame. The output boxes by CPN were input for the CEN. The actions for CPN network included 4 translation actions (Up, Down, Left, Right), 4 scaling action (Horizontal shrink, Vertical shrink, Horizontal zoom, Vertical zoom), and 1 terminal action (Stop), and the reward corresponded to the change in IOU across an action. For the network architecture, a Fully-Convolutional DenseNet56 \cite{jegou2017one} was used as a backbone along with DQN as the agent for CPN and down-sampling followed by up-sampling architecture for CEN.

Unsupervised video object segmentation is an intuitive task in the computer vision field. A DRL method for this task was proposed by \cite{goel2018unsupervised}, where the authors proposed a motion-oriented unsupervised method for image segmentation in videos (MOREL). They proposed a two-step process to achieve the task in which first a representation of input is learned to understand all moving objects through unsupervised video object segmentation, Then the weights are transferred to the RL framework to jointly train segmentation network along with policy and value function. The first part of the method takes two consecutive frames as input and predicts a number of segmentation masks, corresponding object translations, and camera translations. They used a modified version of actor-critic \cite{mnih2016asynchronous}\cite{schulman2017proximal}\cite{van2016deep} for the network of first step. Following the unsupervised fashion, the authors used the approach similar to \cite{vijayanarasimhan2017sfm} and trained the network to interpolate between consecutive frames and used the masks and translations to estimate the optical flow using the method that was proposed in Spatial Transformer Networks \cite{jaderberg2015spatial}. They also used structural dissimilarity (DSSIM) \cite{wang2004image} to calculate reconstruction loss and actor-critic \cite{mnih2016asynchronous} algorithm to learn policy in the second step.

A DRL method for dynamic semantic face video segmentation was proposed by \cite{wang2020dynamic}, where Deep Feature Flow \cite{zhu2017deep} was utilized as the feature propagation framework and RL was used for an efficient and effective scheduling policy. The method involved dividing frames into key ($I_k$) and non-key ($I_i$), and using the last key frame features for performing segmentation of non-key frame. The actions made by the policy network corresponded to categorizing a frame as $I_k$ or $I_i$ and the state consisted of deviation information and expert information, where the deviation information described the difference between current $I_i$ and last $I_k$ and expert information encapsulated the key decision history. The authors utilized FlowNet2-s model \cite{ilg2017flownet} as an optical flow estimation function, and divided the network into feature extraction module and task-specific module. After policy network which consisted of one convolution layer, 4 fully connected layers and 2 concatenated channels consisting of KAR (Key all ratio: Ratio between key frame and every other frame in decision history) and LKD (Last key distance: Distance between current and last key frame) predicted the action, If the current frame is categorized as key frame the feature extraction module produced the frame features and task-specific module predicted the segmentation, However if the frame is categorized as a non-key frame the features from the last key frame along with the optical flow was used by the task-specific module to predict the segmentation. The authors proposed two types of reward functions, The first reward function was calculated by considering the difference between the IOU for key and non-key actions. The second reward function was proposed for a situation when ground truth was not available and was calculated by considering the accuracy score between segmentation for key and non-key actions.

\begin{spacing}{1.2}
\footnotesize
\begin{longtable}{|p{0.09\linewidth}|p{0.04\linewidth}|p{0.10\linewidth}|p{0.09\linewidth}|p{0.15\linewidth}|p{0.09\linewidth}|p{0.15\linewidth}|p{0.13\linewidth}|} 
\caption{Comparing various methods associated with video. First group for video object segmentation, second group for action recognition and third group for video summarisation} 

\label{tab:vid}
\\
\hline
Approaches &
  Year &
  Training Technique &
  Actions &
  Remarks &
  Backbone &
  Performance &
  Datasets and Source code \\ \hline \hline
  Object segmentation in videos\cite{sahba2016deep} &
  2016 &
   Deep Belief Network \cite{chen2015spectral} &
   Actions changed local values in sub-images &
   States: sub-images. Reward: quality of segmentation. &
  Not specified &
  Not specified &
  Not specified \\ \hline
  Surgical gesture segmentation and classification \cite{liu2018deep} &
  2018 &
   Trust Region Policy Optimization (TRPO) \cite{schulman2015trust} &
   2 types: optimal step size and gesture class &
   States: TCN [\cite{lea2017temporal}, \cite{leal2016learning}] and future frames. Reward: encourage larger steps and minimize action errors. Statistical language model \cite{richard2016temporal} for gesture probability. &
  Spacial CNN \cite{lea2016segmental} &
   Comparable accuracy, and higher edit and F1 scores as compared to methods like SD-SDL \cite{sefati2015learning}, Bidir LSTM \cite{dipietro2016recognizing}, LC-SC-CRF \cite{lea2016learning}, Seg-ST-CNN \cite{lea2016segmental}, TCN \cite{lea2016temporal}, etc &
   JIGSAWS [\cite{ahmidi2017dataset}, \cite{gao2014jhu}] benchmark dataset \newline \href{https://github.com/Finspire13/RL-Surgical-Gesture-Segmentation}{\textcolor{blue}{\underline{Available Code}}}\\ \hline
  Cutting agent for video object segmentation \cite{han2018reinforcement} &
  2018 &
  DQN &
   8 actions: 4 translation actions (Up, Down, Left, Right), 4 scaling action (Horizontal shrink, Vertical shrink, Horizontal zoom, Vertical zoom) and 1 terminal action (Stop) &
   States: input frame, action history and segmentation mask. Reward: change in IOU. cutting-policy network for box-context pair and cutting-execution network for mask generation &
  DenseNet \cite{jegou2017one} &
   Higher mean region similarity, counter accuracy and temporal stability \cite{perazzi2016benchmark} as compared to methods like MSK \cite{perazzi2017learning}, ARP \cite{jun2017primary}, CTN \cite{jang2017online}, VPN \cite{jampani2017video}, etc. &
   DAVIS dataset \cite{perazzi2016benchmark} and the YouTube Objects dataset \cite{jain2014supervoxel}, \cite{prest2012learning} \\ \hline
 Unsupervised video object segmentation (MOREL) \cite{goel2018unsupervised} &
  2018 &
  Actor-critic (a2c) \cite{mnih2016asynchronous} &
  Not specified &
  States: consecutive frames. Two step process with optical flow using Spatial Transformer Networks \cite{jaderberg2015spatial} and reconstruction loss using structural dissimilarity \cite{wang2004image}. &
  Multi-layer CNN &
 Higher total episodic reward as compared to methods that used actor-critic without MOREL &
  59 Atari games. \newline \href{https://github.com/vik-goel/MOREL}{\textcolor{blue}{\underline{Available Code}}}\\ \hline
  
 Face video segmentation \cite{wang2020dynamic} &
  2020 &
  Not specified &
  2 actions: categorising a frame as a key or a non-key &
  States: deviation information which described the difference between current non-key and last key decision, and expert information which encapsulated the key decision history. Reward: improvement in mean IOU/accuracy score between segmentation of key and non-key frames  &
  Multi-layer CNN & Higher mean IOU then other methods like  DVSNet \cite{xu2018dynamic}, DFF \cite{zhu2017deep}. &
  300VW dataset \cite{shen2015first} and Cityscape dataset \cite{cordts2016cityscapes} \\ \hline

 Multi-agent Video Object Segmentation \cite{vecchio2020mask} &
  2020 &
  DQN &
  Actions of 2 types: movement actions (up, down, left and right) and set action (action to place location prior at a random location on the patch) &
   States: input frame, optical flow \cite{ilg2017flownet} from previous frame and action history. Reward: clicks generated by gamification. Down-sampling and up-sampling similar to U-Net \cite{ronneberger2015u} &
  DenseNet \cite{huang2017densely} &
  Higher mean region similarity and contour accuracy \cite{perazzi2016benchmark} as compared to semi-supervised  methods such as SeamSeg \cite{avinash2014seamseg}, BSVS \cite{marki2016bilateral}, VSOF \cite{tsai2016video}, OSVOS \cite{caelles2017one} and weakly-supervised methods such as GVOS \cite{spampinato2016gamifying}, Spftn \cite{zhang2017spftn} &
  DAVIS-17 dataset \cite{perazzi2016benchmark} \\ \hline \hline
 Skeleton-based Action Recognition  \cite{tang2018deep} &
  2018 &
  DQN &
  3 actions: shifting to left, staying the same and shifting to right &
  States: Global video information and selected frames. Reward: change in  categorical probability. 2 step network (FDNet) to filter frames and GCNN for action labels &
  Multi-layer CNN &
  Higher cross subject and cross view metrics for NTU+RGBD dataset \cite{shahroudy2016ntu}, and higher accuracy for SYSU-3D  \cite{hu2015jointly} and  UT-Kinect Dataset \cite{xia2012view} when compared with other methods like Dynamic Skeletons \cite{hu2015jointly}, HBRNN-L \cite{du2015hierarchical}, Part-aware LSTM \cite{shahroudy2016ntu}, LieNet-3Blocks \cite{huang2017deep}, Two-Stream CNN \cite{li2018co},  etc. &
   NTU+RGBD \cite{shahroudy2016ntu}, SYSU-3D  \cite{hu2015jointly} and UT-Kinect Dataset  \cite{xia2012view} \\ \hline \hline
 Video summarisation \cite{zhou2018deep} &
  2018 &
  DQN &
  2 actions: selecting and rejecting the frame &
  tates: bidirectional LSTM \cite{huang2015bidirectional} produced states by input frame features. Reward: Diversity-Representativeness Reward  Function. &
  GoogLeNet \cite{szegedy2015going} & Higher F-score \cite{zhang2016video} as compared to methods like Uniform sampling, K-medoids, Dictionary selection \cite{elhamifar2012see}, Video-MMR \cite{li2010multi}, Vsumm \cite{de2011vsumm}, etc. &
  TVSum \cite{song2015tvsum} and SumMe \cite{gygli2014creating}. \newline \href{https://github.com/KaiyangZhou/pytorch-vsumm-reinforce}{\textcolor{blue}{\underline{Available Code}}} \\ \hline
 Video summarization \cite{zhou2018video} &
  2018 &
  Duel DQN Double DQN &
  2 actions: selecting and rejecting the frame &
  States: sequence of frames Reward: Diversity-Representativeness Reward Function 2 stage implementation: classification and summarisation network using bidirectional GRU network and LSTM \cite{huang2015bidirectional} &
  GoogLeNet \cite{szegedy2015going} & Higher F-score \cite{zhang2016video} as compared to methods like Dictionary selection \cite{elhamifar2012see}, GAN \cite{mahasseni2017unsupervised}, DR-DSN \cite{zhou2018deep}, Backprop-Grad \cite{panda2017weakly}, etc in most cases. &
  TVSum \cite{song2015tvsum} and CoSum \cite{chu2015video} datasets. \newline \href{https://github.com/KaiyangZhou}{\textcolor{blue}{\underline{Available Code}}} \\ \hline
  
  Video summarization in Ultrasound \cite{liu2020ultrasound} &
  2020 &
  Not specified &
  2 actions: selecting and rejecting the frame &
  States: frame latent scores Reward: $R_{det}$, $R_{rep}$ and $R_{div}$ bidirectional LSTM \cite{huang2015bidirectional} and Kernel temporal segmentation \cite{potapov2014category} &
  Not specified & Higher F1-scores in supervised and unsupervised fashion as compared to methods like FCSN \cite{rochan2018video} and DR-DSN \cite{zhou2018deep}. &
  Fetal Ultrasound \cite{kirwan2010nhs} \\ \hline


\end{longtable}
\end{spacing}

Video object segmentation using human-provided location priors have been capable of producing promising results. An RL method for this task was proposed by \cite{vecchio2020mask}, in which the authors proposed  MASK-RL, a multiagent RL framework for object segmentation in videos. They proposed a weakly supervised method where the location priors were provided by the user in form of clicks using gamification (Web game to collect location priors by different users) to support the segmentation and used a Gaussian filter to emphasize the areas. The segmentation network is fed a 12 channel input tensor that contained a sequence of video frames and their corresponding location priors (3 $\times$ 3 color channels + three gray-scale images). The authors used a fully convoluted DenseNet \cite{huang2017densely} with down-sampling and up-sampling similar to U-Net \cite{ronneberger2015u} and an LSTM \cite{hochreiter1997long} for the segmentation network. For the RL method, the actor takes a series of steps over a frame divided into a grid of equal size patches and makes the decision whether there is an object in the patch or not. In their MDP formulation the states consisted of the input frame, optical flow (computed by \cite{ilg2017flownet}) from the previous frame, patch from the previous iteration, and the episode location history, the actions consisted of movement actions (up, down, left and right) and set action (action to place location prior at a random location on the patch), and two types of rewards one for set actions and one for movement actions were used. The reward was calculated using the clicks generated by the game player.

Action recognition is an important task in the computer vision field which focuses on categorizing the action that is being performed in the video frame. To address the problem a deep progressive RL (DPRL) method for action recognition in skeleton-based videos was proposed by \cite{tang2018deep}. The authors proposed a method that distills the most informative frames and discards ambiguous frames by considering the quality of the frame and the relationship of the frame with the complete video along with a graph-based structure to map the human body in form of joints and vertices. DPRL was utilized to filter out informative frames in a video and graph-based CNNs were used to learn the spatial dependency between the joints. The approach consisted of two sub-networks, a frame distillation network (FDNet) to filter a fixed number of frames from input sequence using DPRL and GCNN to recognize the action labels using output in form of a graphical structure by the FDNet. The authors modeled the problem as an MDP where the state consisted of the concatenation of two tensors $F$ and $M$, where $F$ consisted of global information about the video and $M$ consisted of the frames that were filtered, The actions which correspond to the output of FDNet were divided into three types: shifting to left, staying the same and shifting to the right, and the reward function corresponded to the change in probability of categorizing the video equal to the ground truth clipped it between [-1 and 1] and is provided by GCNN to FDNet. 

Video summarization is a useful yet difficult task in the computer vision field that involves predicting the object or the task that is being performed in a video. A DRL method for unsupervised video summarisation was proposed by \cite{zhou2018deep}, in which the authors proposed a Diversity-Representativeness reward system and a deep summarisation network (DSN) which was capable of predicting a probability for each video frame that specified the likeliness of selecting the frame and then take actions to form video summaries. They used an encode-decoder framework for the DSN where GoogLeNet \cite{szegedy2015going} pre-trained on ImageNet \cite{russakovsky2015imagenet} \cite{deng2009imagenet} was used as an encoder and a bidirectional RNNs (BiRNNs) topped with a fully connected (FC) layer was used as a decoder. The authors modeled the problem as an MDP where the action corresponded to the task of selecting or rejecting a frame. They proposed a novel Diversity-Representativeness Reward Function in their implementation, where diversity reward corresponded to the degree of dissimilarity among the selected frames in feature space, and representativeness reward measured how well the generated summary can represent the original video. For the RNN unit they used an LSTM \cite{hochreiter1997long} to capture long-term video dependencies and used REINFORCE algorithm for training the policy function. 

\begin{figure}[!h]
	\centering \includegraphics[width=1.0\columnwidth]{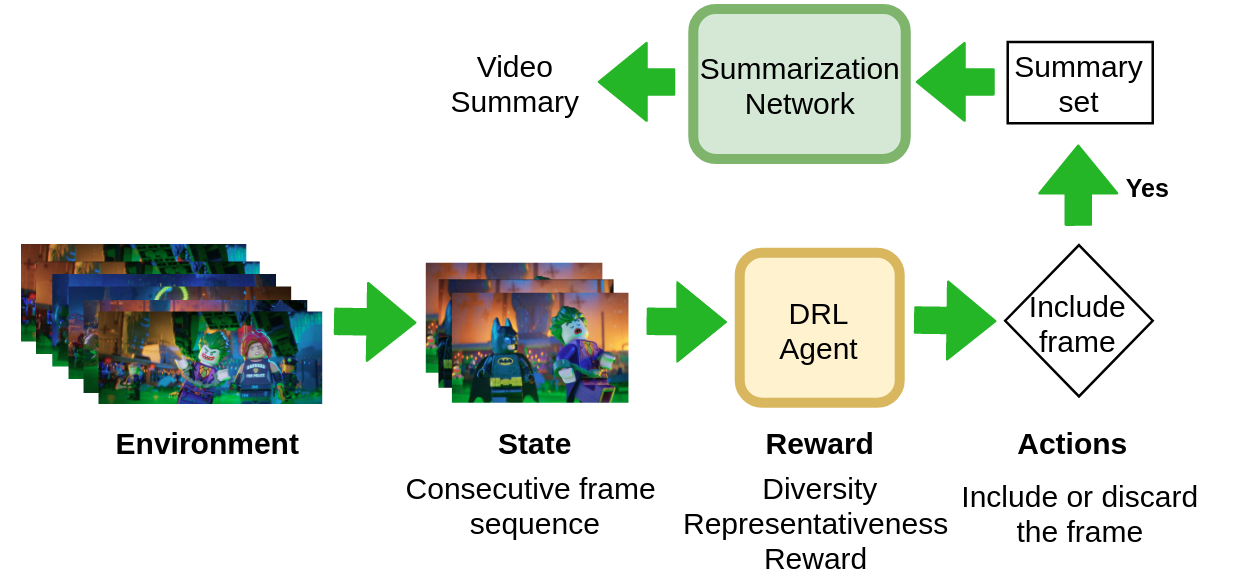}
		\caption{DRL implementation for video summarization. For state a sequence of consecutive frames are used and the DRL agent decided whether to include the frame in the summary set that is used to predict video summary.}
	\label{fig:summary}
\end{figure}

An improvement to \cite{zhou2018deep} was proposed by \cite{zhou2018video}, where the summarisation network was implemented using Deep Q-learning (DQSN), and a trained classification network was used to provide a reward for training the DQSN. The approach included using (Bi-GRU) bidirectional recurrent networks with a gated recurrent unit (GRU) \cite{cho2014learning} for both classification and summarisation network. The authors first trained the classification network using a supervised classification loss and then used the classification network with fixed weights for the classification of summaries generated by the summarisation network. The summarisation network included an MDP-based framework in which states consisted of a sequence of video frames and actions reflected the task of either keeping the frame or discarding it. They used a structure similar to Duel-DQN where value function and advantage function are trained together. In their implementation, the authors considered 3 different rewards: Global Recognisability reward using the classification network with +1 as reward and -5 as punishment, Local Relative Importance Reward for rewarding the action of accepting or rejecting a frame by summarisation network, and an Unsupervised Reward that is computed globally using the unsupervised diversity-representativeness (DR) reward proposed in \cite{zhou2018deep}. The authors trained both the networks using the features generated by GoogLeNet \cite{szegedy2015going} pre-trained on ImageNet \cite{deng2009imagenet}. 

A method for video summarization in Ultrasound using DRL was proposed by \cite{liu2020ultrasound}, in which the authors proposed a deep summarisation network in an encoder-decoder fashion and used a bidirectional LSTM (Bi-LSTM) \cite{huang2015bidirectional} for sequential modeling. In their implementation, the encoder-decoder convolution network extracted features from video frames and fed them into the Bi-LSTM. The RL network accepted states in form of latent scores from Bi-LSTM and produced actions, where the actions consist of the task of including or discarding the video frame inside the summary set that is used to produce video summaries. The authors used three different rewards $R_{det}$, $R_{rep}$ and $R_{div}$ where $R_{det}$ evaluated the likelihood of a frame being a standard diagnostic plane, $R_{rep}$ defined the representativeness reward and $R_{div}$ was the diversity reward that evaluated the quality of the selected summary. They used Kernel temporal segmentation (KTS) \cite{potapov2014category} for video summary generalization.

Various works associated with video analysis have been summarised and compared in Table \ref{tab:vid} and a basic implementation of video summarization using DRL has been shown in Fig. \ref{fig:summary}, where the states consist of a sequence of video frames. The DRL agent performs actions to include or discard a frame from the summary set that is later used by the summarization network to predict video summary. Each research paper propose their own reward function for this application, for example \cite{zhou2018deep} and \cite{zhou2018video} used diversity representativeness reward function and \cite{liu2020ultrasound} used a combination of various reward functions.

\section{Others  Applications}
\label{sec:otherapplications}
Object manipulation refers to the task of handling and manipulating an object using a robot. A method for deformable object manipulation using RL was proposed by \cite{matas2018sim}, where the authors used a modified version of Deep Deterministic Policy Gradients (DDPG) \cite{lillicrap2015continuous}. They used the simulator Pybullet \cite{coumans2016pybullet} for the environment where the observation consisted of a $84 \times 84 \times 3$ image, the state consists of joint angles and gripper positions and action of four dimensions: first three for velocity and lasts for gripper velocity was used. The authors used sparse reward for the task that returns the reward at the completion of the task. They used the algorithm to perform tasks such as folding and hanging cloth and got a success rate of up to 90\%.

Visual perception-based control refers to the task of controlling robotic systems using a visual input. A virtual to real method for control using semantic segmentation was proposed by \cite{hong2018virtual}, in which the authors combined various modules such as, Perception module, control policy module, and a visual guidance module to perform the task. For the perception module, the authors directly used models such as DeepLab \cite{chen2017deeplab} and ICNet \cite{zhao2018icnet}, pre-trained on ADE20K \cite{zhou2017scene} and  Cityscape \cite{cordts2016cityscapes}, and used the output of these model as the state for the control policy module. They implemented the control policy module using the actor-critic \cite{mnih2016asynchronous} framework, where the action consisted of forward, turn right, and turn left. In their implementation, a reward of 0.001 is given at each time step. They used the Unity3D engine for the environment and got higher success and lower collision rate than other implementations such as ResNet-A3C and Depth-A3C.  

Automatic tracing of structures such as axons and blood vessels is an important yet challenging task in the field of biomedical imaging. A DRL method for sub-pixel neural tracking was proposed by \cite{dai2019deep}, where the authors used 2D grey-scale images as the environment. They considered a full resolution 11px $\times$ 11px window and a 21px $\times$ 21px window down-scaled to 11px $\times$ 11px as state and the actions were responsible for moving the position of agent in 2D space using continuous control for sub-pixel tracking because axons can be smaller then a pixel. The authors used a reward that was calculated using the average integral of intensity between the agent’s current and next location, and the agent was penalized if it does not move or changes directions more than once. They used an Actor-critic \cite{mnih2016asynchronous} framework to estimate value and policy functions.

An RL method for automatic diagnosis of acute appendicitis in abdominal CT images was proposed by \cite{al2019reinforcement}, in which the authors used RL to find the location of the appendix and then used a CNN classifier to find the likelihood of Acute Appendicitis, finally they defined a region of low-entropy (RLE) using the spatial representation of output scores to obtain optimal diagnosis scores. The authors considered the problem of appendix localization as an MDP, where the state consisted of a $50 \times 50 \times 50$ volume around the predicted appendix location, 6 actions (2 per axis) were used and the reward consisted of the change in distance between the predicted appendix location and actual appendix location across an action. They utilized an Actor-critic \cite{mnih2016asynchronous} framework to estimate policy and value functions.
\begin{spacing}{1.2}
\footnotesize
\begin{longtable}{|p{0.10\linewidth}|p{0.04\linewidth}|p{0.10\linewidth}|p{0.1\linewidth}|p{0.15\linewidth}|p{0.1\linewidth}|p{0.15\linewidth}|p{0.1\linewidth}|} 
\caption{Comparing various other methods besides landmark detection, object detection, object tracking, image registration, image segmentation, video analysis, that is associated with DRL} 

\label{tab:oth}
\\
\hline
Approaches &
  Year &
  Training Technique &
  Actions &
  Remarks &
  Backbone &
  Performance &
  \shortstack{Datasets \\ Source code}\\ \hline \hline
  
  Object manipulation \cite{matas2018sim} &
  2018 &
  Rainbow DDPG &
  4 actions: 3 for velocity 1 for gripper velocity &
  State: joint angle and gripper position. Reward: at the end of task. &
  Multi layer CNN &
  Success rate up to 90\% &
  Pybullet \cite{coumans2016pybullet}. 
  \newline \href{https://github.com/JanMatas/Rainbow_ddpg}{\textcolor{blue}{\underline{Code}}}
\\ \hline

  Visual based control \cite{hong2018virtual} &
  2018 &
  Actor-critic (a3c) \cite{mnih2016asynchronous} &
  3 actions: forward, turn right and turn left &
  State: output by backbones. Reward: 0.001 at each time-step. &
  DeepLab \cite{chen2017deeplab} and ICNet \cite{zhao2018icnet} &
  Higher success and lower  collision rate then ResNet-A3C and Depth-A3C &
  Unity3D engine \\ \hline
  
  Automatic tracing  \cite{dai2019deep} &
  2019 &
  Actor-critic \cite{mnih2016asynchronous} &
  4 actions &
  State: 11px $\times$ 11px window. Reward: average integral of intensity between the agent’s current and next location. &
  Multi layer CNN &
  Comparable convergence $\%$ and average error as compared to other methods like Vaa3D software \cite{peng2010v3d} and APP2 neuron tracer \cite{xiao2013app2} & Synthetic and microscopy dataset \cite{bass2017detection} \\ \hline
  
  Automatic diagnosis (RLE) \cite{al2019reinforcement} &
  2019 &
  Actor-critic \cite{mnih2016asynchronous} & 6 actions: 2 per axis &
  State: $50 \times 50 \times 50$ volume. Reward: change in distance error. &
  Fully connected CNN &
  Higher sensitivity and specificity as compared to only CNN classifier and CNN classifier with RL without RLE. & Abdominal CT Scans \\ \hline
  
  Learning to paint \cite{huang2019learning} &
  2019 & Actor-critic with DDPG &
  Actions control the stoke parameter: location, shape, color and transparency &
  State: Reference image, Drawing canvas and time step. Reward: change in discriminator score (calculated by WGAN-GP \cite{gulrajani2017improved} across an action.  GANs \cite{goodfellow2014generative} to improve image quality &
  ResNet18 \cite{he2016deep} &
  Able to replicate the original images to a large extent, and better resemblance to the original image as compared to SPIRAL \cite{ganin2018synthesizing} with same number of brush strokes. &
  MNIST \cite{lecun1998mnist}, SVHN \cite{netzer2011reading}, CelebA \cite{liu2015deep} and ImageNet \cite{russakovsky2015imagenet}. 
  \newline \href{https://github.com/hzwer/ICCV2019-LearningToPaint}{\textcolor{blue}{\underline{Code}}}
  \\ \hline
  
  Guiding medical robots \cite{hase2020ultrasound} &
  2020 & Double-DQN, Duel-DQN &
  5 actions: up, down, left, right and stop &
  State: probe position. Reward: Move closer: 0.05, Move away: -0.1, Correct stop: 1.0, Incorrect stop: -0.25. &
  ResNet18 \cite{he2016deep} &
  Higher \% of policy correctness and reachability as compared to CNN Classifier, where MS-DQN showed the best results &
  Ultrasound Images \href{https://github.com/hhase/sacrumdata-set}{Dataset}. 
  \newline \href{https://github.com/hhase/spinal-navigation-rl}{\textcolor{blue}{\underline{Code}}}
  \\ \hline
  
  Crowd counting \cite{Liu2020WeighingCS} &
  2020 & DQN &
  9 actions: -10, -5, -2, -1, +1, +2, +5, +10 and end &
  State: weight vector $W_{t}$ and image feature vector $FV_{I}$. Reward: Intermediate reward and ending reward &
  VGG16 \cite{simonyan2014very} &
  Lower/comparable mean squared error (MSE) and mean absolute error (MAE) as compared to other methods like DRSAN \cite{liu2018crowd}, PGCNet \cite{yan2019perspective}, MBTTBF \cite{sindagi2019multi}, S-DCNet \cite{xiong2019open}, CAN \cite{liu2019context}, etc.  &
  The ShanghaiTech (SHT) Dataset \cite{zhang2016single}, The UCFCC50 Dataset \cite{idrees2013multi} and The UCF-QNRF Dataset \cite{idrees2018composition}. 
  \newline \href{https://github.com/poppinace/libranet}{\textcolor{blue}{\underline{Code}}}
  \\ \hline

Automated Exposure bracketing \cite{wang2020learning}&
  2020 & Not Specified &
  selecting optimal bracketing from candidates &
  State: quality of generated HDR image. Reward: improvement in peak signal to noise ratio &
  AlexNet \cite{krizhevsky2017imagenet} &
  Higher peak signal to noise ratio as compared to other methods like Barakat \cite{barakat2008minimal}, Pourreza-Shahri \cite{pourreza2015exposure}, Beek \cite{van2018improved}, etc.  &
  Proposed benchmark dataset.
  \newline \href{https://github.com/wzhouxiff/EBSNetMEFNet}{\textcolor{blue}{\underline{Code/data}}}
  \\ \hline

 Urban Autonomous driving \cite{toromanoff2020end} &
  2020 & Rainbow-IQN &
  36 or 108 actions: ($9\times4$) or ($27\times4$), 9/27 steering and 4 throttle &
  State: environment variables like traffic light, pedestrians, position with respect to center lane. Reward: generated by CARLA waypoint API &
  Resnet18 \cite{he2016deep} &
  Won the 2019 camera only CARLA challenge \cite{ros2019carla}.  & CARLA urban driving simulator \cite{ros2019carla}
  \newline \href{https://github.com/valeoai/learningbycheating}{\textcolor{blue}{\underline{Code}}}

  \\ \hline
  
 Mitigating bias in Facial Recognition \cite{wang2020mitigating} &
  2020 & DQN &
  3 actions:(Margin adjustment) staying the same, shifting to a larger value and shifting to a smaller value &
  State: the race group, current adaptive margin and bias  between the race group and Caucasians. Reward: change in the sum of inter-class and intra-class bias &
  Multi-layer CNN &
  Proposed algorithm had higher verification accuracy as compared to other methods such as CosFace \cite{wang2018cosface} and ArcFace \cite{deng2019arcface}.  &  RFW \cite{wang2019racial} and proposed novel datasets: BUPT-Globalface and BUPT-Balancedface
\newline \href{http://www.whdeng.cn/RFW/index.html}{\textcolor{blue}{\underline{Data}}}
  \\ \hline
  
  Attention mechanism to improve CNN performance \cite{Qifeng_ECCV2020} & 2020 & DQN \cite{mnih2015human} & Actions are weights for every location or channel in the feature map. & State: Feature map at each intermediate layer of model. Reward: predicted by a LSTM model. & ResNet-101 \cite{he2016deep} & Improves the performances of \cite{SENet_CVPR2018}, \cite{SRM_ICCV2019} and \cite{CBAM_ECCV2018}, which attend on feature channel, spatial-channel and style, respectively & ImageNet \cite{deng2009imagenet} \\ \hline

\end{longtable}
\end{spacing}

\begin{figure}[!h]
	\centering \includegraphics[width=1.0\columnwidth]{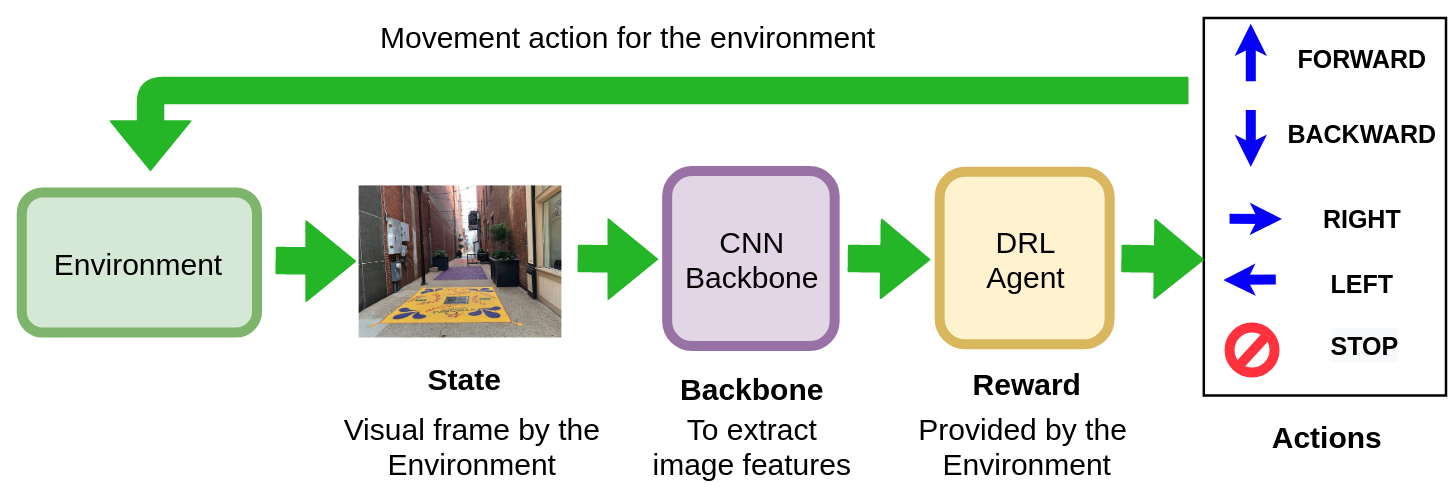}
		\caption{A general DRL implementation for agent movement with visual inputs. The state is provided by the environment based on which the agent performs movement actions to get a new state and a reward from the environment.}
	\label{fig:move}
\end{figure}
Painting using an algorithm is a fantastic yet challenging task in the computer vision field. An automated painting method was proposed by \cite{huang2019learning}, where the authors introduced a model-based DRL technique for this task. The specified work involved using a neural renderer in DRL, where the agent was responsible for making a decision about the position and color of each stroke, and making long-term decisions to organize those strokes into a visual masterpiece. In this work, GANs \cite{goodfellow2014generative} were employed to improve image quality at pixel-level and DDPG \cite{lillicrap2015continuous} was utilized for determining the policy. The authors formulated the problem as an MDP, where the state consisted of three parts: the target image $I$, the canvas on which actions (paint strokes) are performed $C_t$, and the time step. The actions corresponding to a set of parameters that controlled the position, shape, color, and transparency of strokes, and for reward the WGAN with gradient penalty (WGAN-GP) \cite{gulrajani2017improved} was used to calculate the discriminator score between the target image $I$ and the canvas $C_t$, and the change in discriminator score across an action (time-step) was used as the reward. The agent that predicted the stroke parameters was trained in actor-critic \cite{mnih2016asynchronous} fashion with backbone similar to Resnet18 \cite{he2016deep}, and the stroke parameters by the actor were used by the neural renderer network to predict paint strokes. The network structure of the neural renderer and discriminator consisted of multiple convolutions and fully connected blocks.   

A method for guiding medical robots using Ultrasound images with the help of DRL was proposed by \cite{hase2020ultrasound}. The authors treated the problem as an MDP where the agent takes the Ultrasound images as input and estimates the state hence the problem became Partially observable MDP (POMDP). They used Double-DQN and Duel-DQN for estimating Q-Values and ResNet18 \cite{he2016deep} backbone for extracting feature to be used by the algorithm along with Prioritized Replay Memory. In their implementation the action space consisted of 8 actions (up, down, left, right, and stop), probe position as compared to the sacrum was used as the state and the reward was calculated by considering the agent position as compared to the target (Move closer: 0.05, Move away: -0.1, Correct stop: 1.0, Incorrect stop: -0.25). In their implementation, the authors proposed various architectures such as V-DQN, M-DQN, and MS-DQN for the task and performed experimentation on Ultrasound images.

Crowd counting is considered a tricky task in computer vision and is even trickier for humans. A DRL method for crowd counting was proposed by \cite{Liu2020WeighingCS}, where the authors used sequential decision making to approach the task through RL. In the specified work, the authors proposed a DQN agent (LibraNet) based on the motivation of a weighing scale. In their implementation crowd counting was modeled using a weighing scale where the agent was responsible for adding weights on one side of the scale sequentially to balance the crowded image on the other side. The problem of adding weights on one side of the pan for balancing was formulated as an MDP, where state consisted weight vector $W_{t}$ and image feature vector $FV_{I}$, and the actions space was defined similar to scale weighing and money system \cite{van2001optimal} containing values $(-10, -5, -2, -1, +1, +2, +5, +10, end)$. For reinforcing the agent two different rewards: ending reward and intermediate reward were utilized, where ending reward (following \cite{caicedo2015active}) was calculated by comparing the absolute value error between the ground-truth count and the accumulated value with the error tolerance, and three counting specific rewards: force ending reward, guiding reward and squeezing reward were calculated for the intermediate rewards.

Exposure bracketing is a method used in digital photography, where one scene is captured using multiple exposures for getting a high dynamic range (HDR) image. An RL method for automated bracketing selection was proposed by \cite{wang2020learning}. For flexible automated bracketing selection, an exposure bracketing selection network (EBSNet) was proposed for selecting optimal exposure bracketing and a multi-exposure fusion network (MEFNet) for generating an HDR image from selected exposure bracketing which consisted of 3 images. Since there is no ground truth for the exposure bracketing selection procedure, an RL scheme was utilized to train the agent (EBSNet). The authors also introduced a novel dataset consisting of a single auto-exposure image that was used as input to the EBSNet, 10 images with varying exposures from which EBSNet generated probability distribution for 120 possible candidate exposure bracketing ($C^3_{10}$) and a reference HDR image. The reward for EBSNet was defined as the difference between peak signal-to-noise ratio between generated and reference HDR for the current and previous iteration, and the MEFNet was trained by minimizing the Charbonnier loss \cite{barron2019general}. For performing the action of bracketing selection ESBNet consisted of a semantic branch using AlexNet \cite{krizhevsky2017imagenet} for feature extraction, an illumination branch to understand the global and local illuminations by calculating a histogram of input and feeding it to CNN layers, and a policy module to generate a probability distribution for the candidate exposure bracketing from semantic and illumination branches. The neural network for MEFNet was derived from HDRNet \cite{gharbi2017deep}.

Autonomous driving in an urban environment is a challenging task, because of a large number of environmental variables and constraints. A DRL approach to this problem was proposed by \cite{toromanoff2020end}. In their implementation, the authors proposed an end-to-end model-free RL method, where they introduced a novel technique called Implicit Affordances. For the environment, the CARLA Simulator \cite{dosovitskiy2017carla} was utilized, which provided the observations and the training reward was obtained by using the CARLA waypoint API. In the novel implicit affordances technique the training was broken into two phases, The first phase included using a Resnet18 \cite{he2016deep} encoder to predict the state of various environment variables such as traffic light, pedestrians, position with respect to the center lane, etc., and the output features were used as a state for the RL agent, For which a modified version of Rainbow-IQN Ape-X \cite{hessel2017rainbow} was used. CARLA simulator accepts actions in form of continuous steering and throttle values, so to make it work with Rainbow-IQN which supports discrete actions, the authors sampled steering values into 9 or 27 discrete values and throttle into 4 discrete values (including braking), making a total of 36($9\times4$) or 108($27\times4$) actions.

Racial discrimination has been one of the hottest topics of the 21st century. To mitigate racial discrimination in facial recognition, \cite{wang2020mitigating} proposed a facial recognition method using skewness-aware RL. According to the authors, the reason for racial bias in facial recognition algorithms can be either due to the data or due to the algorithm, so the authors provided two ethnicity-aware datasets, BUPT-Globalface and BUPT-Balancedface along with an RL based race balanced network (RL-RBN). In their implementation, the authors formulated an MDP for adaptive margin policy learning where the state consisted of three parts: the race group (0: Indian, 1: Asian, 2: African), current adaptive margin, and bias or the skewness between the race group and Caucasians. A DQN was used as a policy network that performed three actions (staying the same, shifting to a larger value, and shifting to a smaller value) to change the adaptive margin, and accepted reward in form of change in the sum of inter-class and intra-class bias.

Attention mechanisms are currently gaining popularity because of their powerful ability in eliminating uninformative parts of the input to leverage the other parts having a more useful information. Recently, attention mechanism has been integrated into typical CNN models at every individual layer to strengthen the intermediate outputs of each layer, in turn improving the final predictions for recognition in images. This model is usually trained with a weakly supervised method, however, this optimization method may lead to sub-optimal weights in the attention module. Hence, \cite{Qifeng_ECCV2020} proposed to train attention module by deep Q-learning with an LSTM model is trained to predict the reward, the whole process is called Deep REinforced Attention Learning (DREAL).

Various works specified here have been summarised and compared in Table \ref{tab:oth} and general implementation of a DRL method to control an agents movement in an environment has been shown in fig \ref{fig:move} where state consists of an image frame provided by the environment, the DRL agent predicts actions to move the agent in the environment providing next state and the reward is provided by the environment, for example, \cite{hong2018virtual}.

\section{Future Perspectives}
\label{sec:future}
\subsection{Challenge Discussion}
DRL is a powerful framework, which has been successfully applied to various computer vision applications including landmark detection, object detection, object tracking, image registration, image segmentation, video analysis, and other computer vision applications. DRL has also demonstrated to be an effective alternative for solving difficult optimization problems, including tuning parameters, selecting augmentation strategies, and neural architecture search (NAS). However, most approaches, that we have reviewed, assume a stationary environment, from which observations are made. Take landmark detection as an instance, the environment takes into account the image itself, and each state is defined as an image patch consisting of the landmark location. In such a case, the environment is known while the RL/DRL framework naturally accommodates a dynamic environment, that is the environment itself evolves with the state and action. Realizing the full potential of DRL for computer vision requires solving several challenges. In this section, we would like to discuss the challenges of DRL in computer vision for real-world systems.
\begin{itemize}
    \item \textbf{Reward function:} In most real-world applications, it is hard to define a specified reward function because it requires the knowledge from different domains that may not always be available. Thus, the intermediate rewards at each time step are not always easily computed. Furthermore, a reward function with too long delay will make training difficult. In contrast, assigning a reward for each action requires careful and manual human design.
    \item \textbf{Continuous state and action space:} Training an RL system on a continuous state and action space is challenging because most RL algorithms, i.e. Q learning, can only deal with discrete states and discrete action space. To address this limitation, most existing works discretize the continuous state and action space.
    \item \textbf{High-dimensional state and action space}: Training Q-function on a high-dimensional action space is challenging. For this reason, existing works use low-dimensional parameterization, whose dimensions are typically less than 10 with an exception \cite{krebs2017robust} that uses 15-D and 25-D to model 2D and 3D registration, respectively.
    \item \textbf{Environment is complicated:} Almost all real-world systems, where we would want to deploy DRL/RL, are partially observable and non-stationary. Currently, the approaches we have reviewed assume a stationary environment, from which observations are made. However, the DRL/RL framework naturally accommodates dynamic environment, that is the environment itself evolves with the state and action. Furthermore, those systems are often stochastic and noisy (action delay, sensor and action noise) as compared to most simulated environments.
    \item\textbf{Training data requirement:} RL/DRL requires a large amount of training data or expert demonstrations. Large-scale datasets with annotations are expensive and hard to come by. 
    
\end{itemize}

More details of challenges that embody difficulties to deploy RL/DRL in the real world are discussed in \cite{challenges}. In this work, they designed a set of experiments and analyzed their effects on common RL agents. Open-sourcing an environmental suite, \href{https://github.com/google-research/realworldrl_suite}{realworldrl-suite} \cite{realworldRL} is provided in this work as well.

\subsection{DRL Recent Advances}
Some advanced DRL approaches such as Inverse DRL, Multi-agent DRL, Meta DRL, and imitation learning are worth the attention and may promote new insights for many machine learning and computer vision tasks.
\begin{itemize}
    \item \textbf{Inverse DRL:} DRL has been successfully applied into domains where the reward function is clearly defined. However, this is limited in real-world applications because it requires knowledge from different domains that may not always be available. Inverse DRL is one of the special cases of imitation learning. An example is autonomous driving, the reward function should be based on all factors such as driver's behavior, gas consumption, time, speed, safety, driving quality, etc. In real-world scenario, it is exhausting and hard to control all these factors. Different from DRL, inverse DRL \cite{inverseRL_2020}, \cite{inverseDRL}, \cite{IRL_2020_CVPR}, \cite{IRL_2} a specific form of imitation learning \cite{imiation_learning}, infers the reward function of an agent, given its policy or observed behavior, thereby avoiding a manual specification of its reward function. In the same problem of autonomous driving, inverse RL first uses a dataset collected from the human-generated driving and then approximates the reward function. Inverse RL has been successfully applied to many domains \cite{inverseDRL}. Recently, to analyze complex human movement and control high-dimensional robot systems, \cite{online_IRL} proposed an online inverse RL algorithm. \cite{RL_IRL} combined both RL and Inverse RL to address planning problems in autonomous driving. 
    \item \textbf{Multi-Agent DRL:} Most of the successful DRL applications such as game\cite{multi-agent-game}, \cite{multi-agent-game2}, robotics\cite{multi-agent-robotics}, and autonomous driving \cite{multi-agent-autonomous}, stock trading \cite{MARL_stock}, social science \cite{MARL_social}, etc., involve multiple players that requires a model with multi-agent. Take autonomous driving as an instance, multi-agent DRL addresses the sequential decision-making problem which involves many autonomous agents, each of which aims to optimize its own utility return by interacting with the environment and other agents \cite{MARL_autonomous}. Learning in a multi-agent scenario is more difficult than a single-agent scenario because non-stationarity \cite{multi_nonstationarity}, multi-dimensionality \cite{MARL_autonomous}, credit assignment \cite{MARL_assignment}, etc., depend on the multi-agent DRL approach of either fully cooperative or fully competitive. The agents can either collaborate to optimize a long-term utility or compete so that the utility is summed to zero. Recent work on Multi-Agent RL pays attention to learning new criteria or new setup \cite{new_multi_RL}. 
    \item \textbf{Meta DRL:} As aforementioned, DRL algorithms consume large amounts of experience in order to learn an individual task and are unable to generalize the learned policy to newer problems. To alleviate the data challenge, Meta-RL algorithms \cite{schweighofer2003meta}, \cite{meta_RL} are studied to enable agents to learn new skills from small amounts of experience. Recently, there is a research interest in meta RL~\cite{nagabandi2018learning}, \cite{gupta2018meta}, \cite{saemundsson2018meta}, \cite{rakelly2019efficient}, \cite{liu2019taming}, each using a different approach. For benchmarking and evaluation of meta RL algorithms, \cite{yu2020meta} presented Meta-world, which is an open-source simulator consisting of 50 distinct robotic manipulation tasks.
    \item \textbf{Imitation Learning:} Imitation learning is close to learning from demonstrations which aims at training a policy to mimic an expert's behavior given the samples collected from that expert. Imitation learning is also considered as an alternative to RL/DRL to solve sequential decision-making problems. Besides inverse DRL, an imitation learning approach as aforementioned, behavior cloning is another imitation learning approach to train policy under supervised learning manner. Bradly et al. \cite{imitation_1} presented a method for unsupervised third-person imitation learning to observe how other humans perform and infer the task. Building on top of Deep Deterministic Policy Gradients and Hindsight Experience Replay, Nair et al. \cite{clone_DRL} proposed behavior cloning Loss to increase imitating the demonstrations. Besides Q-learning, Generative Adversarial Imitation Learning \cite{gail_IL} proposes P-GAIL that integrates imitation learning into the policy gradient framework. P-GAIL considers both smoothness and causal entropy in policy update by utilizing Deep P-Network \cite{P_Dnn}.
    \end{itemize}

\section*{Conclusion}
Deep Reinforcement Learning (DRL) is nowadays the most popular technique for an artificial agent to learn closely optimal strategies by experiences. This paper aims to provide a state-of-the-art comprehensive survey of DRL applications to a variety of decision-making problems in the area of computer vision. In this work, we firstly provided a structured summarization of the theoretical foundations in Deep Learning (DL) including AutoEncoder (AE), Multi-Layer Perceptron (MLP), Convolutional Neural Network (CNN), and Recurrent Neural Network (RNN). We then continued to introduce key techniques in RL research including model-based methods (value functions, transaction models, policy search, return functions) and model-free methods (value-based, policy-based, and actor-critic). Main techniques in DRL were thirdly presented under two categories of model-based and model-free approaches. We fourthly surveyed the broad-ranging applications of DRL methods in solving problems affecting areas of computer vision, from landmark detection, object detection, object tracking, image registration, image segmentation, video analysis, and many other applications in the computer vision area. We finally discussed several challenges ahead of us in order to realize the full potential of DRL for computer vision. Some latest advanced DRL techniques were included in the last discussion.

\newpage

\vskip 0.2in
\bibliographystyle{plain}
\bibliography{arxiv.bib}
\end{document}